\documentclass{ecai} 

\usepackage{latexsym}
\usepackage{amssymb}
\usepackage{amsmath}
\usepackage{amsthm}
\usepackage{booktabs}
\usepackage{enumitem}
\usepackage{graphicx}
\usepackage{color}
\usepackage{xcolor}
\usepackage{multirow}
\usepackage{caption}

\usepackage{listings}
\lstdefinestyle{pythoncode}{
    language=Python,
    backgroundcolor=\color[HTML]{FFFFFF}, % White background
    commentstyle=\color[HTML]{008000}, % Darker green comments
    keywordstyle=\color[HTML]{0000FF}, % Blue keywords
    stringstyle=\color[HTML]{A31515}, % Dark red strings
    basicstyle=\ttfamily\color[HTML]{000000}\footnotesize, % Black text
    captionpos=t,
    keepspaces=true,
    showspaces=false,
    showstringspaces=false,
    showtabs=false,
    tabsize=4,
    frame=tb,
    rulecolor=\color[HTML]{000000}, % Frame color (same as background for invisibility)
    morekeywords={as,assert,nonlocal,with,yield,self,True,False,None}, % Adding more Python keywords
    identifierstyle=\color[HTML]{000000}, % Default text color
    rulesepcolor=\color[HTML]{FFFFFF}, % Color between line numbers and code (same as background)
}

\newcommand{\BibTeX}{B\kern-.05em{\sc i\kern-.025em b}\kern-.08em\TeX}

\begin{document}

\begin{frontmatter}

%%% Use this command to specify your submission number.
%%% In doubleblind mode, it will be printed on the first page.

\paperid{552} 

\title{NeSIG: A Neuro-Symbolic Method for Learning to Generate Planning Problems}

%%% Use this combinations of commands to specify all authors of your 
%%% paper. Use \fnms{} and \snm{} to indicate everyone's first names 
%%% and surname. This will help the publisher with indexing the 
%%% proceedings. Please use a reasonable approximation in case your 
%%% name does not neatly split into "first names" and "surname".
%%% Specifying your ORCID digital identifier is optional. 
%%% Use the \thanks{} command to indicate one or more corresponding 
%%% authors and their email address(es). If so desired, you can specify
%%% author contributions using the \footnote{} command.

\author[A,B]{\fnms{Carlos}~\snm{Núñez-Molina}\orcid{0000-0003-1450-7323}\thanks{Corresponding Author. Email: ccaarlos@ugr.es.}}
\author[A,B]{\fnms{Pablo}~\snm{Mesejo}\orcid{0000-0001-9955-2101}}
\author[A,B]{\fnms{Juan}~\snm{Fernández-Olivares}\orcid{0000-0002-7391-882X}}

\address[A]{University of Granada, Spain}
\address[B]{Andalusian Institute of Data Science and Computational Intelligence (DaSCI)}

\begin{abstract}
    In the field of Automated Planning there is often the need for a set of planning problems from a particular domain, e.g., to be used as training data for Machine Learning or as benchmarks in planning competitions. In most cases, these problems are created either by hand or by a domain-specific generator, putting a burden on the human designers. In this paper we propose NeSIG, to the best of our knowledge the first domain-independent method for automatically generating planning problems that are valid, diverse and difficult to solve. We formulate problem generation as a Markov Decision Process and train two generative policies with Deep Reinforcement Learning to generate problems with the desired properties. We conduct experiments on three classical domains, comparing our approach against handcrafted, domain-specific instance generators and various ablations. Results show NeSIG is able to automatically generate valid and diverse problems of much greater difficulty (15.5 times more on geometric average) than domain-specific generators, while simultaneously reducing human effort when compared to them. Additionally, it can generalize to larger problems than those seen during training.
    %Results show NeSIG is able to automatically generate valid and diverse problems of much greater difficulty than the alternative approaches, thus reducing human effort when compared to domain-specific generators.
\end{abstract}

\end{frontmatter}

\section{Introduction}

%\textcolor{red}{Debo mencionar cómo se generan los problemas por encima (partiendo de un init state, añadiendo objetos y átomos y entonces ejecutando acciones para generar el goal), ya que en la sección 4.1.1 me baso en esto para hablar de la consistencia.}
%Max 7 pags. + 1 para referencias
Automated Planning (AP) \cite{ghallab2016automated} is a subfield of
Artificial Intelligence devoted to providing goal-oriented, deliberative behaviour to both physical and virtual agents, e.g. robots and video game automated players. An automated planner receives as input the description of the planning task to solve, containing the environment dynamics, initial state and goal. It then carries out a search process in order to find a plan (sequence of actions) which achieves the task goal starting from its initial state. Planning tasks are usually described in a declarative, first-order logic (FOL) language such as PDDL (Planning Domain Definition Language) \cite{haslum2019introduction}. The PDDL description consists of a planning domain, containing the environment dynamics, and a planning problem/instance, containing the initial state and goal to achieve. This encoding allows the reuse of the same planning domain for planning tasks with different initial state and/or goal but which share their environment dynamics.

% of two different items: a planning domain, containing the environment dynamics, and a planning problem/instance, containing the initial state and goal to achieve. This encoding allows the reuse of the same planning domain for planning tasks with different initial state and/or goal but which share their environment dynamics.

% The planning task description is usually split into two different items: a planning domain, containing the environment dynamics, and a planning problem/instance, containing the initial state and goal to achieve. This encoding allows the reuse of the same planning domain for planning tasks with different initial state and/or goal but which share their environment dynamics.

Throughout the years, many works \cite{shen1989rule,pasula2007learning,segura2021discovering} have tried to automatically learn planning domains from data, in order to alleviate the burden on domain designers. Nevertheless, the task of generating planning problems has received less attention. In most cases, they need to be created by hand or produced by hard-coded, domain-specific instance generators, which requires great human effort. Having a large set of planning problems is useful for several reasons. The main one is that many approaches which apply Machine Learning (ML) to AP use planning problems as training data, such as those for learning heuristics \cite{shen2020learning} and hierarchical, HTN domains \cite{hogg2008htn}. Additionally, problems are used as benchmarks in planning competitions \cite{vallati20152014} and are useful for domain validation \cite{mccluskey2017engineering}, i.e., ensuring the planning domain faithfully represents the environment dynamics.

% AP -> what it is
% Problem motivation. Why should we care about problem generation -> reduce human effort, useful for ML, etc.

% Problem formulation (what are we trying to achieve)? -> automatically generate problems for some domain, in order to reduce human effort. Problems must be valid, diverse and quality (explain each property briefly).

% Outline NeSIG. -> What it is, what it does, what it needs. Why is it so good.
% First method for generating problems with these three qualities
% Does not require data
% Inputs and outputs
% Consistency language so that consistency rules can be encoded with ease and are interpretable
% Scope limited to STRIPS

% How NeSIG actually works
% Incremental generation (MDP). First generate init, then goal.
% Trained with Deep RL
% NLM
% Experiments and results (we compare with other SOTA) -> conclusions

In this work, we address the problem of automatically generating planning instances for a particular PDDL domain provided by the user,
in order to reduce human effort.
Generated problems must exhibit three desirable properties: \textbf{validity}, \textbf{diversity} and \textbf{quality}.
A problem is considered valid if it is solvable and its initial state is consistent, i.e., it describes a possible initial state of the world.
Diversity refers to the fact that generated problems should all be different from each other.
Finally, the quality of a problem is defined by the user. In this work, we will use difficulty as our only quality measure, i.e., the goal is to generate problems as hard to solve as possible.

% Introducción a nuestro método. Main contribution: generating problems valid, diverse and quality (to the best of our knowledge the first work that generates problems with the three properties for any domain (domain-independent)) for any typed STRIPS domain. Explicación de método. Problem generation as MDP. Neurosymbolic approach: NLM to learn to generate problems with desired properties and policies trained with Deep RL.

% New

As our main contribution, we propose \textbf{NeSIG} (\textit{Neuro-Symbolic Instance Generator}) which, to the best of our knowledge, is the first domain-independent method for automatically generating planning problems that are at the same time valid, diverse and of good quality (i.e., difficult to solve). 
NeSIG receives as inputs the PDDL description of the domain, a set of consistency constraints which generated problems must satisfy, the maximum size of the problems to generate and a list with the predicates and object types that can appear in the problem goals.
Then, it leverages this information to learn to generate valid, diverse and difficult problems for the domain provided as input.
A remarkable feature of our method is that \textbf{it does not need to observe a single example problem} from the domain and, thus,
can be regarded as a \textit{data-free} generative model for planning problems.
Finally, we have limited our scope to typed-STRIPS domains with existential and negative preconditions.\footnote{We have used \textit{Lifted PDDL} as our parser, which only provides support for this subset of ADL.}

Consistency constraints describe the set of properties that problem initial states must satisfy in order to be considered consistent, i.e., so that they describe a possible initial state of the world in the system modeled by the PDDL domain at hand. 
An example consistency constraint would be "\textit{an object cannot be at two places at the same time}".
Since planning problems are often crafted by hand, this consistency information is not encoded in the PDDL domain description. Instead, it is the duty of the human designer to create problems representing possible, \textit{reasonable} situations of the world.
In the case of handcrafted, domain-specific generators,
consistency constraints are hard-coded into their problem-generation procedure,
so that inconsistent problems are never generated.
In other words, consistency information must always be present when generating problems in one way or another.
Therefore, in order to alleviate the burden on human designers,
we propose a novel semi-declarative language that combines FOL and Python-like syntax so that the consistency constraints required by NeSIG can be encoded in an intuitive and interpretable manner,
thus reducing human effort when compared to manual problem design and handcrafted generators.

% OLD

%As our main contribution, we propose \textbf{NeSIG} (\textit{Neuro-Symbolic Instance Generator}) which, to the best of our knowledge, is the first domain-independent method for automatically generating planning problems that are at the same time valid, diverse and of good quality (i.e., difficult to solve). 
%NeSIG receives as inputs the PDDL description of the domain, a set of consistency constraints which generated problems must satisfy, the maximum size of the problems to generate and a list with the predicates and object types which can appear in the problem goals.
%Then, it leverages this information to learn to generate valid, diverse and difficult problems for the domain provided as input.
%A remarkable feature of our method is that \textbf{it does not need to observe a single example problem} from the domain and, thus,
%can be regarded as a \textit{data-free} generative model for planning problems.
%As a side contribution,
%we also propose a novel semi-declarative language that combines FOL and Python-like syntax for encoding the problem consistency constraints in an intuitive and interpretable manner,
%thus reducing human effort even further.
%Finally, we have limited our scope to typed-STRIPS domains with existential and negative preconditions.\footnote{We have used \textit{Lifted PDDL} as our parser, which only provides support for this subset of ADL.} 

We formulate problem generation as a Markov Decision Process (MDP) \cite{sutton2018reinforcement}, in which a problem is generated in a series of steps.
Firstly, the initial state of the problem is generated by successively adding objects and atoms to an initially empty state (or some other state predefined by the user). Then, a sequence of domain actions (i.e., the actions present in the planning domain) are executed from the generated initial state to arrive at a goal state, where a subset of the atoms are selected to form the problem goal, according to the goal types and predicates provided by the user. 
We use Deep Reinforcement Learning (RL) \cite{sutton2018reinforcement} to train two generative policies, one for generating the problem initial state and one for the goal. 
These policies are learned by Neural Logic Machines (NLMs) \cite{dong2019neural}, a neuro-symbolic deep neural network architecture capable of inductive learning and logic reasoning. 

We test our method on three classical planning domains,
\textit{blocksworld}, \textit{logistics} and \textit{sokoban},
comparing the problems generated by NeSIG with those obtained by handcrafted, domain-specific generators and several ablations.
Results show NeSIG obtains valid and diverse problems of much greater difficulty than the alternative approaches (15.5 times more than domain-specific generators on geometric average),
while requiring little human effort,
as our approach removes the need to design the problems manually or program a handcrafted generator.
Additionally, NeSIG exhibits remarkable generalization abilities when tested on larger problems than those seen during training:
in \textit{logistics}, it generates harder problems than the handcrafted generator for up to twice the training size
whereas, in \textit{blocksworld} and \textit{sokoban},
it outperforms the domain-specific generators for all test sizes considered.

% Our results show NeSIG obtains valid problems of greater difficulty than the competitor approaches while maintaining good diversity. Therefore, it provides an automated, domain-independent tool for the generation of valid, diverse and quality planning problems without the need to design the problems manually or program a domain-specific generator, thus reducing human intervention.
% Our method builds on the preliminary work \cite{nunezproposal}.

% The main contribution of this work is NeSIG (\textit{Neuro-Symbolic Instance Generator}), a method which learns to generate planning problems for any given domain\footnote{We limit the scope of this work to typed-STRIPS domains extended with the subset of ADL supported by the \textit{Lifted PDDL} parser (https://github.com/AI-Planning/lifted-pddl).} so that they are valid, diverse and difficult to solve. Our method builds on the preliminary work \cite{nunezproposal}. We formulate problem generation as a Markov Decision Process (MDP) \cite{sutton2018reinforcement}, in which a problem is generated in a series of steps. We use Reinforcement Learning (RL) \cite{sutton2018reinforcement} to train two generative policies, one for generating the initial state of the problem and one for the goal. These policies are learned by Neural Logic Machines (NLMs) \cite{dong2019neural}, a neuro-symbolic deep neural network architecture capable of inductive learning and logic reasoning. To the best of our knowledge, this is the first work which proposes a domain-independent method for generating valid, quality and diverse planning problems.

\section{Related work}

Several works have proposed domain-independent methods for planning problem generation but, to the best of our knowledge, none of them have been able to generate problems that are simultaneously valid, of good quality and diverse. \cite{fern2004learning} proposes a random-walk approach to generate planning problems. It randomly creates an initial state $s_i$ and executes $n$ actions at random to arrive at state $s_g$. Then, it selects a subset of the atoms of $s_g$, which constitutes the goal $g$, and returns the planning problem $(s_i, g)$. Although the problems obtained are always solvable, they may not exhibit the other properties (consistency, quality and diversity), as they are generated at random. \cite{fuentetaja2012planning} also employs a random-walk approach but, unlike the previous work, it uses semantics-related information provided by the user to guarantee the consistency of the problems obtained. Thus, this method always generates valid problems but provides no guarantees about their diversity or quality, since they are also generated at random. \cite{marom2020utilising} follows a different approach. It starts from a predefined goal state and performs a backward search for the initial state. The problems obtained are used to learn a planning heuristic. The proposed method estimates its uncertainty and uses this value to search for problems with the right difficulty for training the heuristic. Hence, this method is able to obtain valid problems of good quality. However, it only works for domains with a single, predefined goal and for which there exist an \textit{inverse transition model}, i.e., for every action $a$ that transitions from state $s$ to $s'$ an inverse action $a'$ that goes from $s'$ to $s$ must exist, which needs to be provided to the method.

Finally, it is worth to mention several works that address a similar problem to the one tackled in this paper. \cite{katz2020generating} proposes a method for obtaining diverse and difficult planning tasks with different causal graphs. This work generates complete tasks (i.e., domain-problem pairs) whereas NeSIG generates planning problems for the particular domain provided by the user. \cite{torralba2021automatic} proposes Autoscale, a method for obtaining valid and diverse problems with graded difficulty for their use in planning competitions. However, unlike our method, Autoscale does not generate problems on its own. Instead, it relies on domain-specific instance generators, selecting a set of problems with graded difficulty among the ones they generate. Therefore, Autoscale can be considered complementary to our approach, as it could be used to select problems among those NeSIG generates. 

%obtaining a series of parameter values which, when input to the generators, result in problems of increasing difficulty (and often also of increasing size).

\section{Background}

%This section briefly explains some background concepts. We first describe how planning tasks are represented and then explain how NLMs work.

\subsection{Planning task representation}

A planning task is a tuple formed by a planning domain and a planning problem (also known as an instance). Both the domain and problem are often represented in a formal FOL-based language such as PDDL \cite{haslum2019introduction}. In PDDL, the domain encodes the existing object types, predicates and actions available to solve the task, detailing for each action its parameters (variables), preconditions (conditions which must be true for the action to be applicable) and effects (how the action modifies the state). This information is encoded in \textit{lifted} form, i.e., in terms of FOL variables which can be instantiated (\textit{grounded}) on objects. On the other hand, the PDDL problem encodes the objects present in that particular instance, the set of atoms which are true at the initial state, and the goal to achieve, represented as a FOL formula (usually as just a conjunction of atoms).

\subsection{Neural Logic Machines}
\label{section:nlm}

A Neural Logic Machine (NLM) \cite{dong2019neural} is a deep neural network capable of learning from FOL data and performing logic reasoning. An NLM receives as input a set of predicates grounded on a set of objects. Then, it sequentially applies first-order rules to obtain a different set of output predicates instantiated on the same objects. Input predicates are represented as binary tensors containing the truth value for each grounding of the predicate on the set of objects. Given some input predicate $p$, if $p(o_i, o_j, o_k)$ is true (where $i,j,k$ represent object indexes), then its associated tensor will contain a value of $1$ at the $(i,j,k)$ position. Output predicates and those inferred internally by the NLM are also represented as tensors, but they contain real values between $0$ and $1$. The NLM operates with these tensors by using neural modules that approximate boolean rules ($and$, $or$, $not$) and quantifications ($\forall$ and $\exists$), being expressive enough to realize a set of Horn clauses. Therefore, NLMs are more expressive than alternative architectures such as Graph Neural Networks \cite{barcelo2020logical}, which is why they are used in this work.

\begin{figure*}[h]
	\centering
	\includegraphics[width=.9\textwidth]{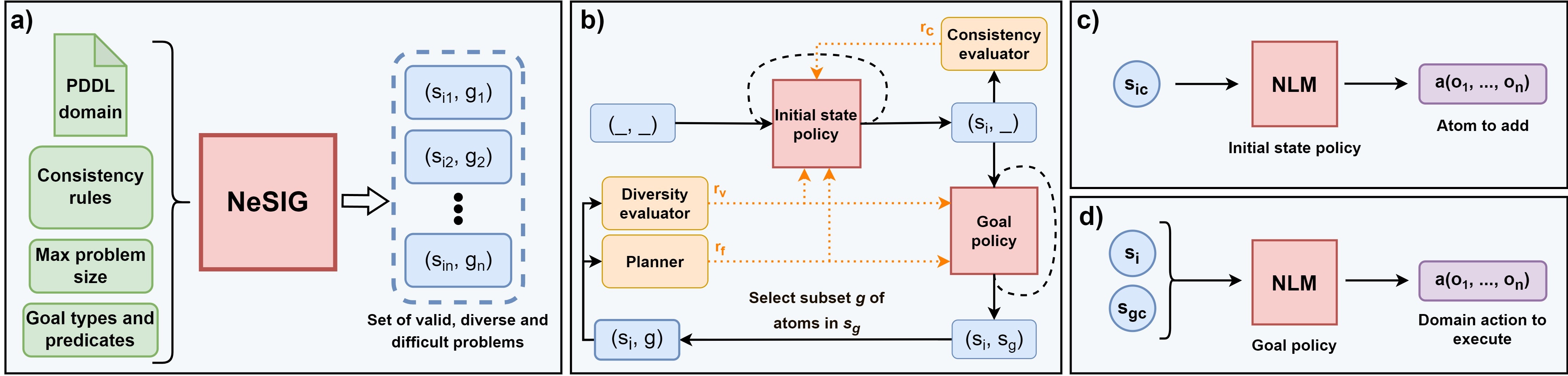}
 \captionsetup{skip=10pt, belowskip=10pt}
	\caption{\textbf{NeSIG.} \textbf{a) Architecture overview.} NeSIG receives as inputs a PDDL domain, several consistency rules and some extra information (maximum problem size and goal types and predicates). It then trains two generative policies with RL (see subfigure b) so that they learn to generate valid, diverse and difficult problems for the domain provided as input. \textbf{b) Policy training with RL.} Dashed lines represent the application of several MDP actions, corresponding to adding an atom to the initial state in the case of the initial state policy (see subfigure c), or executing a domain action in the goal state in the case of the goal policy (see subfigure d). Dotted lines indicate the reward signal, accounting for the consistency $r_c$, diversity $r_v$ and difficulty $r_f$ of the problems generated. \textbf{c) Initial state policy.} It receives an MDP state $(s_{ic}, \_)$ corresponding to a partially-generated initial state and selects the next atom to add to $s_{ic}$. \textbf{d) Goal policy.} It receives an MDP state $(s_i, s_{gc})$ representing a complete initial state but a partially-generated goal state and selects the next domain action to execute in $s_{gc}$.}
    \label{fig:diagram_problem_generation}
\end{figure*}

\section{Neuro-Symbolic instance generation}

% La explicación "informal" de cómo se generan los problemas paso a paso la doy en la sección 4.2 (problem generation as MDP) y quizás también en la introducción

In this section we describe our method, shown in Figure \ref{fig:diagram_problem_generation}. NeSIG takes as inputs a PDDL planning domain, a set of consistency rules generated problems must satisfy and some extra information, corresponding to the maximum size of the problems to generate and a list with the predicates and object types which can appear in the problem goals.\footnote{Additionally, NeSIG may also take as input the list of object types that can be added to the problem initial state during generation. Nonetheless, this is completely optional and is only used for improving NLM efficiency.} It then learns to generate problems for that particular domain so that they are valid, diverse and difficult to solve (see Figure \ref{fig:diagram_problem_generation}a).
Problems are generated via an iterative process that first generates the problem initial state by sequentially adding objects and atoms to some predefined (often empty) state and, then, 
executes domain actions from the initial state generated to arrive at the goal state,
where the problem goal is obtained according to the goal predicates and types specified by the user.
We now discuss how validity, diversity and difficulty are defined and measured, present our novel MDP formulation of problem generation and explain how we leverage Deep RL to learn to generate problems with the desired properties.

% OLD
%We now discuss how these properties are defined and measured,
%present our novel MDP formulation of problem generation in which problems are generated step-by-step,
%and explain how we can learn to generate problems with these desired properties using Deep RL.

\subsection{Problem properties}
\label{section:problem_properties}

%Our main goal is to generate problems for a particular domain which exhibit the properties specified by the user. In this work, we focus on three desirable properties: validity, diversity and difficulty.

\subsubsection{Validity}
\label{section:validity}

% Validity=solvable+consistent
% Solvable by definition, so we focus on consistency
% Consistency definition -> init state that makes sense, that can be in the world
% Not in PDDL, since init state is often generated by a human, so we need to specify it additionally
% Eventual vs consistency rules (due to how initial states are generated incrementally by adding atoms and objects)
% Example of each type of rule in logistics? (reader should know a bit about the domain?)
% Consistency rules can mostly be encoded in logic (put reference)
% Explain our PDDL-Prover language (semi-declarative language)
% Appendix, consistency rules for each domain. Very little human knowledge, a lot less than programming domain-specific generators!!

This property can be decomposed into two sub-properties: \textbf{solvability} and \textbf{consistency}.
A problem is considered solvable if there exists at least one valid plan that achieves the problem goal starting from its initial state, i.e., which \textit{solves} the problem.
By design, every problem generated by NeSIG is solvable,
since the goal of a problem is generated by executing applicable domain actions from its initial state.
A problem is considered \textit{consistent} if its initial state represents a possible initial situation (state) within the system modeled by the planning domain, in other words, if it \textit{makes sense}.
An example consistency rule would be "\textit{an object cannot be at two places at the same time}."
Consistency constraints arise from the semantics of the domain and, as previously explained,
are not encoded in its PDDL description so they need to be provided separately.
Additionally, they depend on human interpretation and preferences.
Going back to our previous example, some user could consider a state where one object is at two different places ($at(o,p_1)$, $at(o,p_2)$) at the same time to be consistent,
and that choice would be completely valid as there is nothing in the PDDL domain that forbids it.
%The reason for this is that planning problems are often generated either by hand or by a human-engineered instance generator,
%so it is the duty of the human designer to ensure generated problems are consistent.

Due to the sequential nature of our proposed method, in which problem initial states are generated by incrementally adding objects and atoms to an initially empty state or some other state provided by the user, we distinguish between \textbf{continuous} and \textbf{eventual} consistency.
A continuous consistency rule is one which must be \textit{continuously} satisfied throughout the entire initial state generation process.
In order to make a continuous-inconsistent state consistent again we would need to remove some atom(s) and/or object(s) from the state, which is forbidden in our method.
For this reason, NeSIG never adds objects or atoms which result in continuous-inconsistent states.
An example continuous consistency rule would be "\textit{an object $o$ cannot be at two places $p_1,p_2$ at the same time}", i.e., $at(o,p_1), at(o,p_2)$ is forbidden. If this constraint is not met, we would need to remove either $at(o,p_1)$ or $at(o,p_2)$ from the initial state which, as previously stated, is forbidden.
On the other hand, eventual consistency rules are those which must be \textit{eventually} satisfied once the initial state has been completely generated, but do not need to be met at each step of the generation process.
An eventual-inconsistent state can be made consistent if some particular combination of object(s) and/or atom(s) are added to it.
Therefore, eventual consistency is only checked at the end of the initial state generation process.
An example eventual consistency rule would be "\textit{the initial state must contain at least one object of type $t$}". If this constraint is not met, we can simply add an object of type $t$ to the state to make it eventual-consistent.

% Rule encoding (semi-declarative language, paper reference, etc.)
% Poner ejemplos para que vean lo fácil que es
% Semi-declarative language -> minimal effort and rules that are easy to understand
% Hablar del consistency evaluator -> métodos para continuous (state+atom+objs) y eventual (final init state)
% Appendix: consistency rules for each domain

These consistency rules are encapsulated in a consistency evaluator that provides two methods.
The first one returns whether the state resulting from adding some atom (and optionally some objects) to the current state is continuous-consistent or not.
The second method receives a completely-generated initial state and checks whether it is eventual-consistent or not.
Although consistency rules must be provided by a human designer on a per-domain basis,
doing so is often much simpler than devising a procedure for generating a diverse set of consistent problems, i.e., programming an instance generator.
To reduce human effort even further, we have designed a novel, semi-declarative language for describing consistency rules.
It allows the construction of first-order logic (FOL) formulas (with counting quantifiers) expressing conditions about state objects and atoms.
For example, the consistency rule "\textit{the initial state must contain at least 3 objects of type $city$}" can be concisely expressed as $TE(x, type(x,city))>=3$, where $TE$ stands for \textit{There Exists} and $x$ is a variable.
These formulas are then automatically evaluated, and their truth value is stored in a Python boolean variable.
Therefore, we can encode consistency rules using either standard Python, FOL or a combination of them.
This choice is transparent to NeSIG and does not impact training.
We provide the consistency rules for each domain in the Appendix, showing how our semi-declarative language makes possible to represent consistency constraints in an interpretable manner with just a few lines of code.

\subsubsection{Diversity}

% No space. Add this to AIJ.
%At first, we tried measuring problem diversity with the approach proposed in \cite{cenamor2019insights}, which extracts a set of planning features from each problem and calculates diversity as the mean distance in terms of these features, although we employed the larger set of features proposed in \cite{fawcett2014improved}.
%Nonetheless, we could not use many of these features because either they could not be extracted for some problems or their value was identical for all problems of the same domain. Additionally, some features explicitly depended on problem difficulty, which was undesirable.

This property measures how different generated problems are from each other. 
In order to measure diversity, we automatically extract a set of interpretable features that describes the objects and atoms of each problem and their relationships.
We say that two objects are \textit{connected} if they are instantiated on the same atom, regardless of position.
Based on this idea, we define the sets of \textit{connection features} $c_\mu$ and $c_\sigma$.
$c_\mu[t_i][p][t_j]$ encodes how many objects of type $t_j$, on average, each object of type $t_i$ is connected to through atoms of predicate type $p$.
Analogously, $c_\sigma[t_i][p][t_j]$ contains the standard deviation instead of the mean number of connections.
For example, a value $c_\mu[city][in][location]=3$ means that each city contains (atom $in$) an average of three locations, whereas $c_\sigma[city][in][location]=2$ means that the standard deviation between the number of locations in each city is 2 (i.e., not every city contains the same number of locations).
In total, we extract 7 groups of features,
corresponding to the number of objects of each type in the problem and, separately for the initial state and goal,
the number of atoms of each predicate type, $c_\mu$ and $c_\sigma$.
They are divided by their sum so that, for each problem, features in each group add up to one.
Then, we calculate the pairwise problem distance as the absolute difference between their feature vectors, dividing distances by $7*2=14$ to normalize them to the $[0,1]$ range.
Finally, the diversity of a problem is equal to its average distance to all the problems in the set (excluding itself).
%a percentage of the closest problems (excluding itself).
%the diversity value of a problem is the average distance with the $N\%$ of closest (most similar) problems to it, excluding itself.
%By focusing on the most similar problems, we significantly penalize the diversity of problems that are very similar (or identical) to several others in the set.

% No space, add to AIJ
%We only use the closest problems when calculating diversity in order to strongly penalize the generation of identical (repeated) problems.
%For instance, if we use $N=20\%$ for a set of 10 problems and, for some problem, its two most similar problems are identical to it, its diversity value will be 0, even if the set contains several other different problems.

\subsubsection{Difficulty}

In this work, we measure the quality of a problem by its difficulty. In other words, our goal is to generate problems which are as hard to solve by a planner as possible (in addition to being consistent and diverse).
We have chosen difficulty as our quality measure because it plays a central role in AP, where great effort has been devoted to studying problem difficulty \cite{cohen2017problem} and developing efficient algorithms for solving difficult problems \cite{bonet2001planning}.
We measure difficulty as the number of nodes a particular planner needed to expand to solve the problem.
Since this measure depends on the planner employed, we calculate problem difficulty with a different set of planners at training and test time, to evaluate whether NeSIG is able to generate problems which are challenging for different planners.

\subsection{Problem generation as MDP}
\label{section:problem_generation_mdp}

We propose to generate problems of the form $(s_i, g)$, where $s_i$ is the problem initial state and $g$ is the goal, via an iterative process which first generates $s_i$ and then $g$.
The initial state generation phase starts either from an empty state (with no objects or atoms) or from some predefined state provided by the user.
Then, at each step, a new atom is added to the initial state and, optionally, one or more new objects.
Once $s_i$ has been completely generated, the goal generation phase begins if the state meets the eventual consistency constraints. Otherwise, the problem is discarded.
%After $s_i$ has been completely generated, the goal generation phase begins.
Starting from $s_i$, the goal generation phase successively executes the actions available in the domain to arrive at another state, known as the goal state $s_g$.
Finally, the goal $g$ is obtained by selecting a subset of the atoms in $s_g$, according to the goal predicates and object types specified by the user. For instance, in the \textit{blocksworld} domain, problem goals only contain atoms of the form \textit{on(block,block)} by design.
This entire process is depicted in Figure \ref{fig:diagram_problem_generation}b and a handcrafted example is provided in the Appendix. It can be formulated as an undiscounted, finite-horizon MDP $(S,A,app,t,r)$:

\begin{itemize}
    \item $S$ is the state space of the MDP. In our case, states correspond to (incomplete or fully-generated) planning problems $s=(s_{ic}, s_{gc})$. We use the subindex $c$ (\textit{current}) to denote when the initial state $s_{ic}$ and goal state $s_{gc}$ may not be completely generated yet.
    
    \item $A$ is the action space, while $app : S \times A \rightarrow \{0,1\}$ is the applicability function which determines if an action can be executed at a state or not. The set of applicable actions $A_{app}$ is different for the initial state and goal generation phases. In the initial state generation phase, $A_{app}$ corresponds to adding a new atom to the initial state $s_{ic}$ which preserves the continuous consistency constraints (see Section \ref{section:validity}). The objects this new atom is instantiated on can already be present in $s_{ic}$ or not. If they are not, we refer to them as \textit{virtual} objects, and are added to $s_{ic}$ alongside their corresponding atom. For example, if the applicable action $add \ ontable(b1)$ is selected and the object $b1$ does not exist in $s_{ic}$, then both the atom $ontable(b1)$ and the object $b1$ will be added to $s_{ic}$. Thus, instantiating atoms on virtual objects is the mechanism we use to add new objects to the problem. In the goal generation phase, $A_{app}$ is the subset of actions in the planning domain for which their preconditions are met at the current goal state $s_{gc}$.
    Additionally, we add a \textit{termination action} $end$ to $A_{app}$. When $end$ is applied during the initial state generation phase, $s_i=s_{ic}$ is fixed and, if $s_i$ is eventual-consistent, the goal generation phase starts. Otherwise, the MDP episode concludes. When $end$ is applied during the goal generation phase, $s_g=s_{gc}$ is fixed, so the problem $(s_i,g)$ is returned and the episode concludes.
    In order to control problem size, we set a maximum number of actions for each generation phase so, if this number is reached, $end$ is executed and the corresponding phase concludes.

    \item $t : S \times A \rightarrow S$ is the transition function. In our setting, $t$ is deterministic and returns the next MDP state (i.e., problem) resulting from executing an applicable action at the current state.

    \item $r: S \times A \rightarrow \mathbb{R}$ is the reward function. In our setting, there are three different reward sub-types accounting for problem consistency, difficulty and diversity.
    At the end of the initial state generation phase, a consistency reward $r_{c}=-1$ is given if $s_i$ is eventual-inconsistent\footnote{Since actions resulting in continuous-inconsistent states are never executed, we know $s_i$ meets the continuous consistency constraints.}, as a form of penalization.
    At the end of the goal generation phase, problems receive a difficulty reward $r_{f}$ equal to the logarithm of their difficulty, and a diversity reward $r_{v}$ equal to their diversity.
    In every other situation, $r_{c}$, $r_{f}$ and $r_{v}$ are all 0.
    Finally, the (total) reward is calculated as follows:
    \begin{equation}
    \label{eq:total_reward}
    r = r_{c} + min\Big( \frac{r_{v}}{\theta}, 1\Big) \cdot r_{f}
    \end{equation}
    where $\theta \in [0,1]$ is a hyperparameter known as the \textit{diversity threshold}.
    We now explain the rationale behind Equation \ref{eq:total_reward}.
    MDP trajectories resulting in eventual-inconsistent problems will receive a reward $r=-1$ in their last sample, since the difficulty and diversity of an inconsistent problem is 0.
    For trajectories resulting in consistent problems, the reward (for the last sample) will be equal to $r_f$ scaled down by a factor $min(r_v/\theta, 1)$, which depends on the diversity:
    if $r_v \geq \theta$, then $r=r_f$ whereas, if $r_v < \theta$, $r_f$ will be scaled down up to a minimum of $r=0$, in case $r_v=0$.
    This reward function $r$ balances problem consistency, diversity and difficulty. By maximizing it, we hope NeSIG will learn to generate consistent problems with a diversity close to $\theta$ (since diversity values $r_v$ larger than $\theta$ do not increase $r$ and values lower than $\theta$ reduce $r$ considerably) and as difficult to solve as possible.
       
    % (since higher diversities $r_v \ge \theta$ do not increase rewards $r$ and lower diversities quickly reduce the reward) and as difficult to solve as possible. 
 
    %\item $r: S \times A \rightarrow \mathbb{R}$ is the reward function, accounting for the validity and quality of the problems. Since all the generated problems are solvable (as $s_g$ is obtained by executing domain actions from $s_i$), we only need to consider the consistency aspect of the validity. To do so, the user must provide a method (consistency validator) to check the consistency of the initial state $s_{ic}$ associated with a given MDP state. It is used to penalize (i.e., give a negative reward) the initial state generation policy for selecting actions resulting in inconsistent states. Once a problem has been completely generated, it is solved with different automated planners. Then, the policies receive a final reward directly proportional to the resolution difficulty of the problem, i.e., the average number of nodes expanded by the planners. 

\end{itemize}

\subsection{Learning to generate problems with RL}
\label{section:learning_problems_RL}

% Two different policies for the initial and goal phases. They direct problem generation towards problems with the desired qualities.
% Each policy as an NLM -> we explain both NLMs/Policies in parallel
% > NLM inputs
%   State: s_ic for init NLM, concatenated s_i and s_gc for goal NLM
%   States represented as tensors which encode the atoms and objects -> see background 3.2
%   For init NLM, we add virtual objects (not neede for goal NLM as no objects are added during this phase)
%   Additional info: percentage of actions (with respect to max) executed in the init/goal phase, number of objects of each type in the problem and the number of atoms of each predicate type in the init state (init NLM) or in the init and the goal states (goal NLM) and the type of each object in the problem

We use two different policies for guiding problem generation. One policy generates the initial state $s_i$ of each problem, whereas the other generates its goal $g$.
Each policy is encoded by a separate NLM (see Section \ref{section:nlm}).

At each step, the corresponding NLM receives information about the current MDP state.
In the case of the initial state policy, it receives a tensor representation of the atoms and objects in the current initial state $s_{ic}$.
This set of objects contains both the actual objects in $s_{ic}$ and the new, virtual objects that can be added to the state alongside the next atom.
The set of virtual objects is automatically inferred from the predicate information encoded in the PDDL domain.
In the case of the goal policy, the NLM receives as input a concatenation of the tensor representations of the initial state $s_i$ and current goal state $s_{gc}$.
Since no new objects can be added during the goal generation phase, no virtual objects are used.
Additionally, both NLMs receive as extra information
the percentage of actions executed in the corresponding phase (relative to the maximum number of actions allowed), for each object its type and whether it is virtual or not, the total number of objects of each type, and the total number of atoms of each predicate type in the initial state and, for the goal policy NLM, also in the goal state. 

% > NLM outputs
%   A real value for each atom, each representing an action. We add a nullary atom representing the termination action END. We mask out non-applicable actions (continuous-inconsistent atoms in init phase or actions whose preconditions are not met in the goal phase), and apply softmax to obtain the probability of applying each action.

The output of the NLM is represented as a new set of atoms, where each atom is associated with a different MDP action $a \in A$, corresponding to either a new atom to add to $s_{ic}$ (for the initial state policy) or a domain action to apply to $s_{gc}$ (for the goal policy),
in addition to the termination action $end$.
The NLM outputs a real value for each atom (action) in this set.
Then, we mask out inapplicable actions $a \notin A_{app}$, corresponding to either atoms that violate the continuous consistency constraints (for the initial state policy) or domain actions whose preconditions are not met at $s_{gc}$ (for the goal policy).
Finally, we apply the softmax function to obtain a probability distribution over applicable actions $a \in App$, from which we sample the action to execute at the current MDP state.

% Policy training with RL
% policies trained with RL to generate problems with difficulty, diversity and consistency.
% We use PPO. We need a critic NLM for each actor NLM. This NLM is identical but, instead of outputing a probability over actions, it outputs the value of the state.
% We tested Generalized Advantage Estimation but found out that the best value of lambda is 1 (corresponding to simply n-step returns).
% We also tested using a policy entropy bonus but found no gain from it (we assume this is because we already motivate diversity with $r_v$).
% Both policies are trained simultaneously in an end-to-end fashion.
% After generating a trajectory, we calculate the undiscounted return. Since, the consistency reward $r_c$ is given to the last sample of the initial generation phase, the goal generation policy does not receive the consistency reward signal (only the difficulty and diversity), whereas the initial policy receives the signal from all these rewards.

In order to train the initial state and goal policies, we resort to the Deep RL algorithm Proximal Policy Optimization (PPO) \cite{schulman2017proximal}.
Since PPO is an actor-critic algorithm, we need to employ an additional critic NLM for each policy, whose sole purpose is to evaluate the current MDP state.
The two policies are trained simultaneously in an end-to-end fashion.
The initial state policy receives rewards accounting for problem consistency, diversity and difficulty.
On the other hand, the reward signal the goal policy receives accounts for diversity and difficulty but not consistency, since the consistency of a problem is independent of its goal $g$ and, thus, of the goal policy.
In order to calculate the PPO advantages, we use the Generalized Advantage Estimation (GAE) \cite{schulman2015high} method. However, we found the best $\lambda$ value to be equal to 1,
which is equivalent to simply calculating advantages with the n-step returns (i.e., not using GAE).
Moreover,
we use a policy entropy bonus as proposed in \cite{schulman2017proximal} to encourage sufficient exploration, in addition to the diversity reward.

%Additionally, we tested a policy entropy bonus as proposed in \cite{schulman2017proximal},
%but found no gains.
%We believe this is due to the fact that we already encourage diversity explicitly with our diversity reward $r_v$.

\section{Experimentation}

In this section we detail our experimental setup and analyze the results of our experiments, where we compare NeSIG against alternative approaches.
%In this section we explain the experiments carried out in this paper. We first detail our experimental setup and then discuss the results of our experiments, where we compare the problems generated by NeSIG with those obtained by alternative approaches, and analyze whether our method can generalize to larger problems than those used during training. 
%\textcolor{red}{TODO}
%We provide our full code and data, including all the generated problems, in the supplementary material.
%Upon paper acceptance, we will provide the link to our public Github repository.
%Additionally, we will also make available a Docker image for easy deployment,
%as we intend for our method to become a staple tool in the AP community.
Our full code and data can be found in GitHub \cite{code}.
We have made available a Docker image for easy deployment, as we intend for our method to become a staple tool in the AP community.

\subsection{Experimental setup}

We perform experiments on a set of diverse and well-known planning domains: \textit{blocksworld}, \textit{logistics} and \textit{sokoban}. In \textit{blocksworld}, a set of stackable blocks needs to be re-assembled with a gripper. \textit{Logistics} represents a transportation task where a set of packages needs to be delivered across locations and cities using airplanes and trucks. \textit{Sokoban} is a challenging puzzle where several boxes must be pushed to their goal locations. In \textit{blocksworld} and \textit{logistics}, the initial state generation state starts from an empty state $s_i$ with no objects or atoms. In \textit{sokoban}, $s_i$ initially describes an empty NxM map with no robots, walls or boxes, which will be added at generation time.
The PDDL description for each domain can be found in the Appendix.

We train NeSIG separately on each domain, performing 5000 training steps using Adam \cite{kingma2014adam} with a learning rate of $10^{-3}$.
Each experiment is run on 25 threads of an AMD EPYC 7742 CPU and one Nvidia A100 GPU, although our method can be trained on consumer-grade GPUs since only 8 GBs of VRAM are needed.
In each training step, we generate 25 problems by executing up to 15 initial state actions (i.e., adding a maximum of 15 atoms to $s_i$) and up to 60 goal actions in \textit{blocksworld} and \textit{logistics}. For \textit{sokoban}, we execute up to 75 goal actions, as this domain is more challenging than the others,
and use a map of size 5x5.
Every 250 training steps, we perform one validation epoch, where 100 problems are generated and the reward $r$ of each problem is obtained using Equation \ref{eq:total_reward}. 
We calculate the \textit{validation score} of the model as the average problem reward and, once training concludes, we load the model checkpoint with the best validation score for testing.
The complete list of hyperparameters is provided in the Appendix. We use almost identical values for each domain so as to show our method needs little hyperparameter tuning.

\begin{figure*}[ht]
  \centering
  \includegraphics[width=0.29\textwidth]{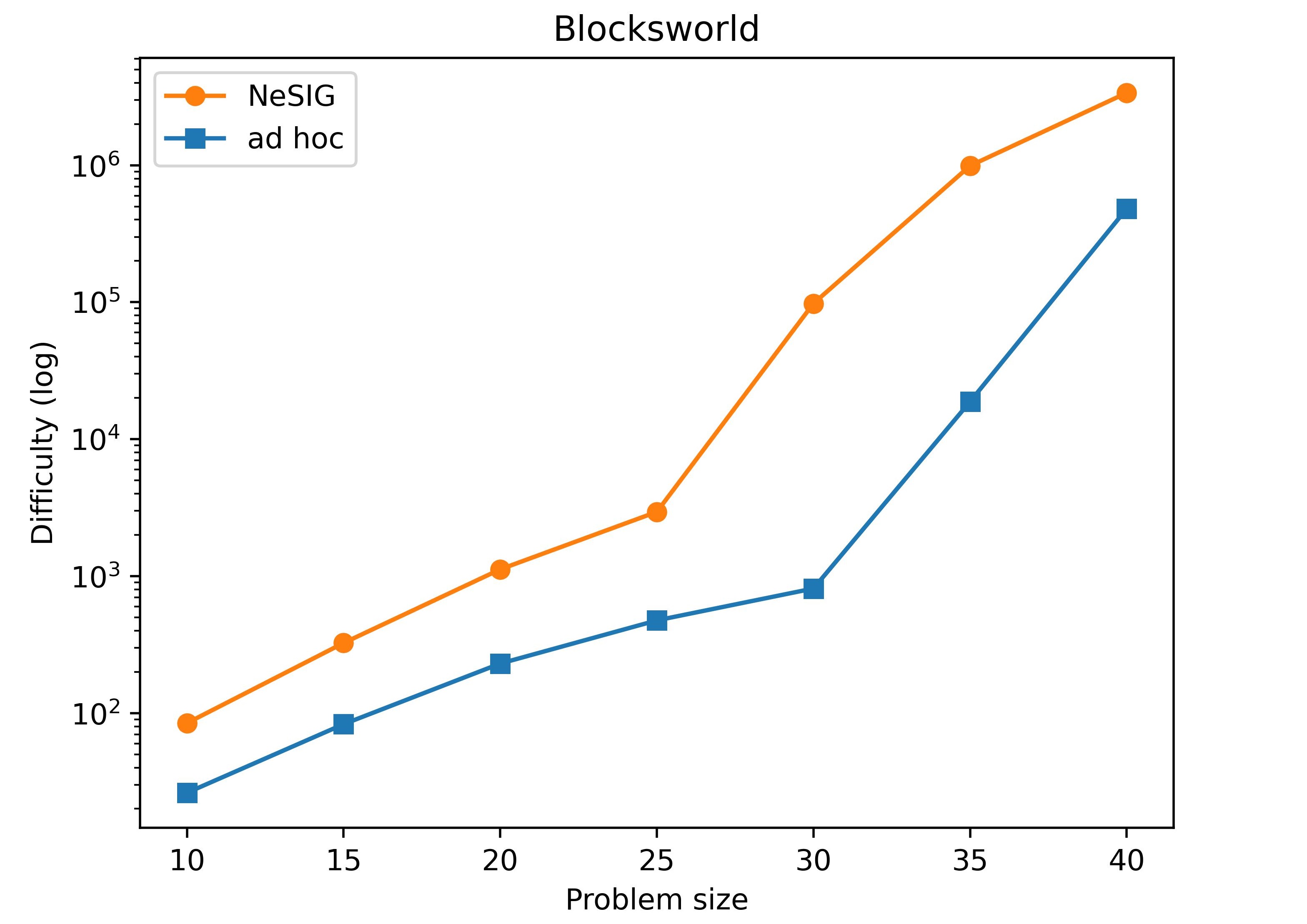}
  \includegraphics[width=0.29\textwidth]{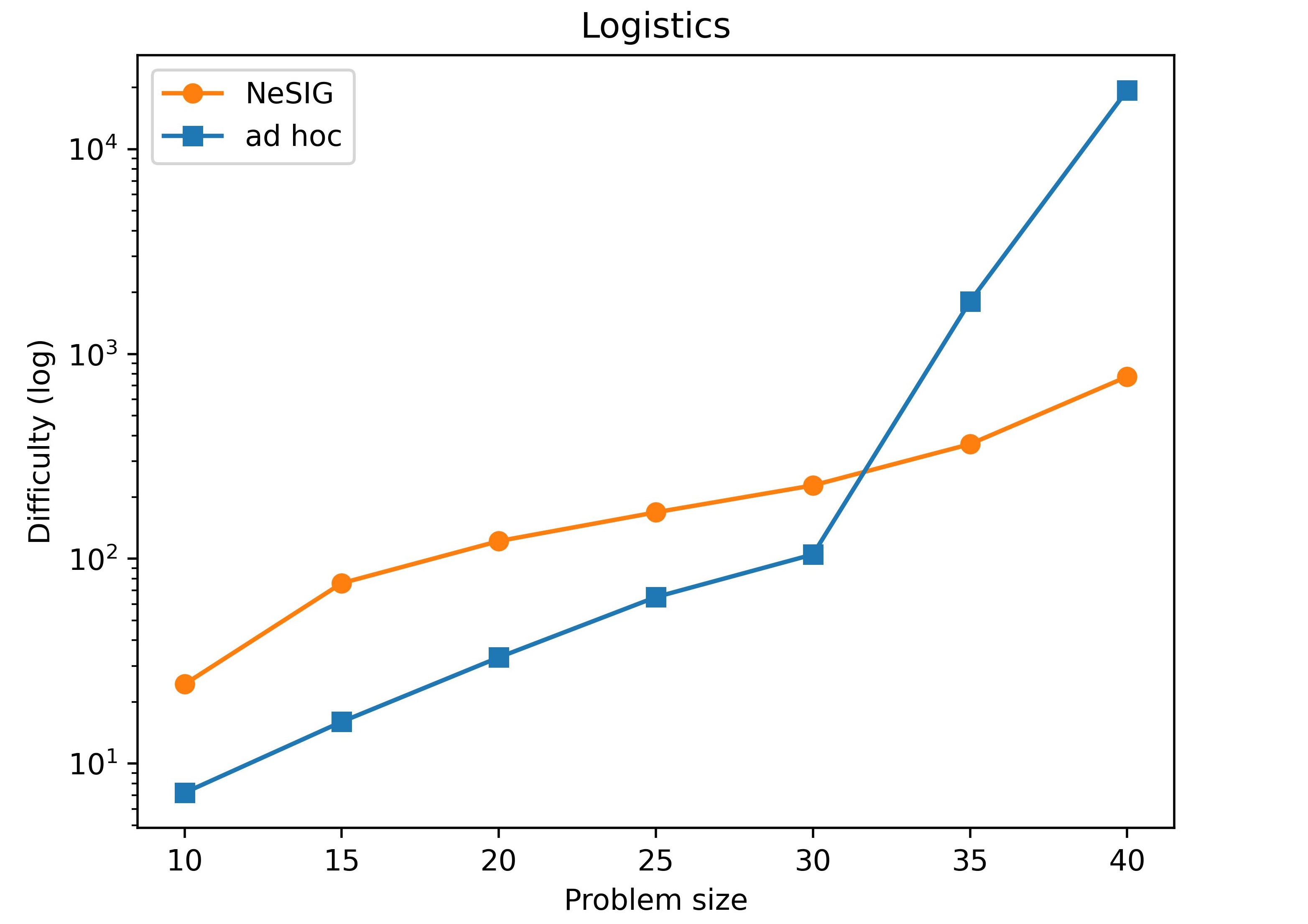}
  \includegraphics[width=0.29\textwidth]{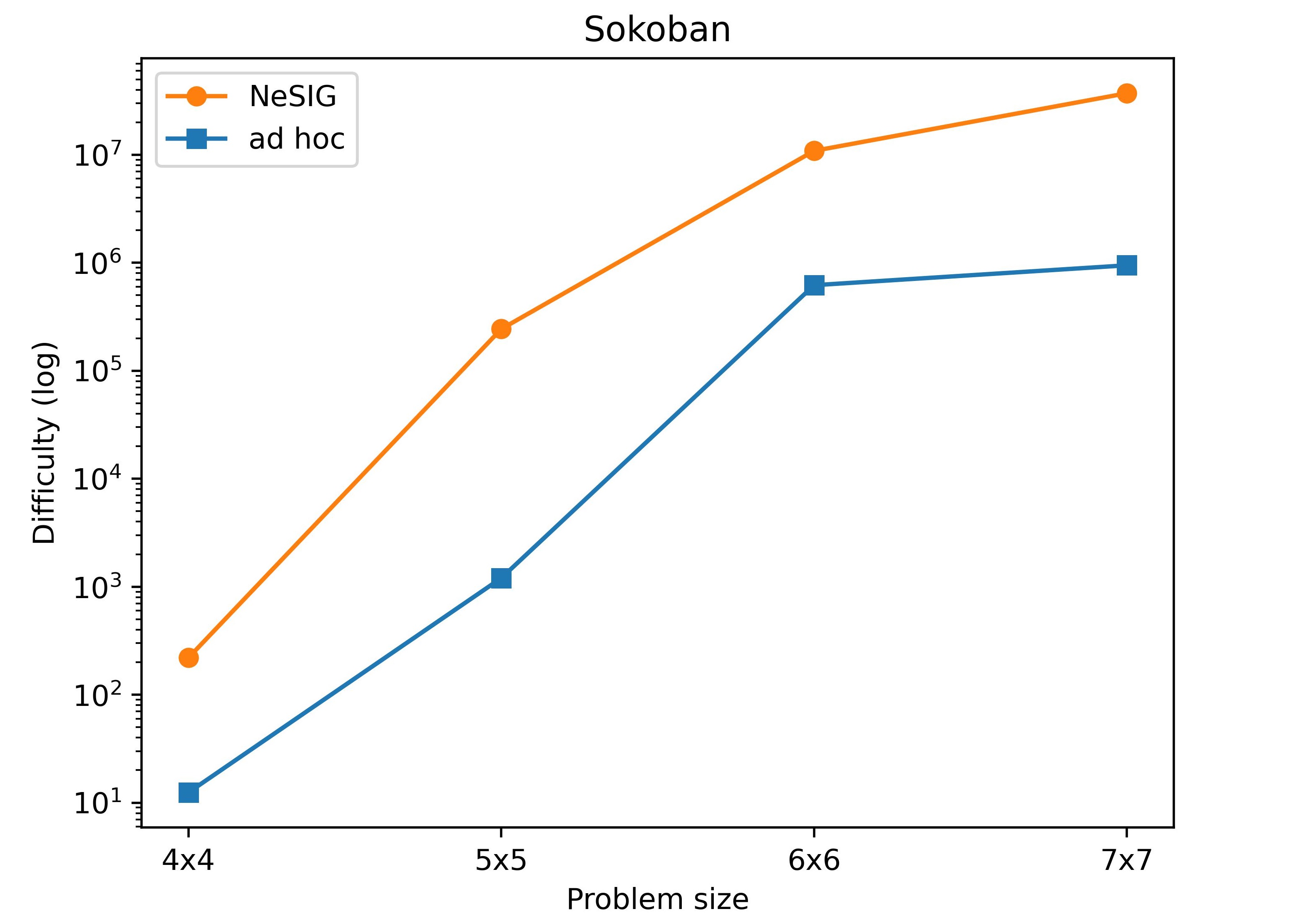}
  \caption{\textbf{Problem size generalization results.}
  The plots show the mean difficulty (in log scale) obtained by NeSIG across five different seeds, when tested on larger (and smaller) problems than those seen during training.
  We also plot the problem difficulty of the domain-specific generators (\textit{ad hoc} models) for comparison purposes.
  In \textit{blocksworld} and \textit{logistics}, problem size is measured as the maximum number of atoms allowed in the initial state $s_i$. In \textit{sokoban}, it is measured by the map size NxM.
  The maximum number of initial state and goal actions used by NeSIG for each problem size, along with the parameters of the \textit{ad hoc} models, are detailed in the Appendix.
  \vspace{0.1cm}
  }
  \label{fig:generalization_plots}
\end{figure*}

Problem difficulty is calculated as the average number of nodes expanded by one or more planners to solve the problem.
We employ the planners provided by FastDownward (FD) \cite{helmert2006fast}.
During training, we solve each problem with LAMA-first \cite{richter2010lama} using up to 500 MB of memory and 5 minutes of planning time, setting a difficulty of $10^6$ for problems that could not be solved under those limits.
At test time, we use LAMA-first, lazy-greedy search with the FF heuristic \cite{hoffmann2001ff} and lazy-greedy with the additive heuristic \cite{bonet2001planning} with a memory limit of 8 GB and time limit of 30 minutes, setting a difficulty of $10^8$ for terminated problems.
We use different planners for training and testing to evaluate whether NeSIG can generate problems that are challenging for several planners.
In the Appendix, we provide experiments with optimal planners.
Finally, for efficiency purposes, we generate small problems during training and then evaluate the generalization abilities of NeSIG by generating larger problems at test time (see Figure \ref{fig:generalization_plots}).

\begin{table*}[ht]
\centering
\caption{\textbf{Same test-size experiment results.} 
The table compares the problems generated by NeSIG, several ablations (\textit{random-init}, \textit{random-goal} and \textit{random-both} models) and the domain-specific generator (\textit{ad hoc} model) in \textit{blocksworld}, \textit{logistics} and \textit{sokoban}.
For each domain and model, we generate 100 test problems with the same maximum size $D$ used during training, corresponding to 15 max atoms in $s_i$.
In \textit{sokoban}, we use a map size of 5x5.
We evaluate the consistency, difficulty, diversity and generation time of the test problems generated, showing for each property its mean value and standard deviation ($\! \pm \!$) across 5 random seeds.
Since the ad hoc models do not require training, we use a single initial random seed (which will be used to deterministically obtain the seed to generate each problem), which is why their std values are always 0.
Consistency is measured as the percentage of problems that meet the eventual consistency rules.
Difficulty is measured as the mean number of nodes the test planners needed to expand to solve the problems.
When calculating the mean difficulty and diversity, we do not consider inconsistent problems.
Time refers to the total generation time (in seconds) needed to generate the whole set of 100 test problems.
}
\label{table:experiments}

% Blocksworld
\begin{tabular}{lccccc}
\toprule
 \multirow{2}{*}{Property} & \multicolumn{5}{c}{Blocksworld} \\
 & NeSIG & random-init & random-goal & random-both & ad hoc \\
 \cmidrule(r){2-6}
 Consistency & $.986 \! \pm \! .01$ & $.14 \! \pm \! .022$ & $.986 \! \pm \! .012$ & $.14 \! \pm \! .022$ & $1.0 \! \pm \! .0$ \\
 Difficulty & $325 \! \pm \! 36$ & $94 \! \pm \! 24$ & $36 \! \pm \! 4$ & $30 \! \pm \! 2$ & $83 \! \pm \! 0$ \\
 Diversity & $.025 \! \pm \! .002$ & $.026 \! \pm \! .014$ & $.033 \! \pm \! .006$ & $.026 \! \pm \! .014$ & $.024 \! \pm \! .0$ \\
 Time & $26 \! \pm \! 4$ & $11 \! \pm \! 2$ & $22 \! \pm \! 1$ & $9 \! \pm \! 0$ & $4 \! \pm \! 0$ \\
\bottomrule
\end{tabular}

\vspace{0.05cm}

% Logistics
\begin{tabular}{lccccc}
\toprule
 \multirow{2}{*}{Property} & \multicolumn{5}{c}{Logistics} \\
 & NeSIG & random-init & random-goal & random-both & ad hoc \\
 \cmidrule(r){2-6}
 Consistency & $.998 \! \pm \! .004$ & $.254 \! \pm \! .047$ & $.994 \! \pm \! .008$ & $.254 \! \pm \! .047$ & $1.0 \! \pm \! .0$ \\
 Difficulty & $76 \! \pm \! 4$ & $14 \! \pm \! 2$ & $13 \! \pm \! 1$ & $5 \! \pm \! 1$ & $16 \! \pm \! 0$ \\
 Diversity & $.196 \! \pm \! .01$ & $.222 \! \pm \! .009$ & $.167 \! \pm \! .003$ & $.25 \! \pm \! .007$ & $.264 \! \pm \! .0$ \\
 Time & $28 \! \pm \! 3$ & $14 \! \pm \! 1$ & $26 \! \pm \! 1$ & $13 \! \pm \! 1$ & $4 \! \pm \! 0$ \\
\bottomrule
\end{tabular}

\vspace{0.05cm}

% Sokoban
\begin{tabular}{lccccc}
\toprule
 \multirow{2}{*}{Property} & \multicolumn{5}{c}{Sokoban} \\
 & NeSIG & random-init & random-goal & random-both & ad hoc \\
 \cmidrule(r){2-6}
 Consistency & $1.0 \! \pm \! .0$ & $.994 \! \pm \! .005$ & $.998 \! \pm \! .004$ & $.994 \! \pm \! .005$ & $1.0 \! \pm \! .0$ \\
 Difficulty & $2.4e5 \! \pm \! 7.6e4$ & $5 \! \pm \! 0$ & $1.8e4 \! \pm \! 1.7e4$ & $6 \! \pm \! 1$ & $1.2e3 \! \pm \! 0$ \\
 Diversity & $.016 \! \pm \! .001$ & $.007 \! \pm \! .0$ & $.013 \! \pm \! .001$ & $.007 \! \pm \! .0$ & $.016 \! \pm \! .0$ \\
 Time & $221 \! \pm \! 17$ & $327 \! \pm \! 8$ & $255 \! \pm \! 36$ & $333 \! \pm \! 4$ & $1019 \! \pm \! 0$ \\
\bottomrule
\end{tabular}
\end{table*}

% Generalization plots

Several methods are compared to NeSIG in our experiments.
First, we employ ablations where either $s_i$ (\textit{random-init} models), $s_g$ (\textit{random-goal} models) or both (\textit{random-both} models) are generated by executing random actions $a \in A_{app}$.
We note our \textit{random-both} model is equivalent to the method proposed in \cite{fuentetaja2012planning}, which also generates $s_i$ and $s_g$ at random.
We do not compare with Autoscale \cite{torralba2021automatic} since it leverages domain-specific generators to obtain problems of graded difficulty, often by gradually incrementing their size,
whereas our goal is instead to maximize problem difficulty given a limit on their size.
For this reason, we directly utilize the ad hoc, domain-specific generators (\textit{ad hoc} models) used in the International Planning Competitions (IPCs) \cite{vallati20152014},
choosing their parameter values to maximize problem diversity (see Appendix for the exact values).
Nonetheless, the \textit{sokoban} generator allowed for little flexibility (e.g., problems of size 5x5 could not have more than two boxes), so we have implemented our own based on a trial and error strategy which obtains $s_i$ by placing objects at random on the grid, randomly moves boxes to obtain $g$, makes sure $g$ can be achieved from $s_i$ and, otherwise, discards the problem and starts again.
For a fair comparison with NeSIG, we discard generated \textit{blocksworld} and \textit{logistics} problems with size smaller than $D-2$, where $D$ is the maximum problem size, measured as the maximum number of atoms in $s_i$.

\subsection{Analysis of results}

% -----

% Paragraphs

% > Same test size experiments
% NeSIG vs ad hoc
% - NeSIG is able to keep similar diversity as ad hoc models. This is remarkable as ad hoc models are random whereas nesig is not
% - Low variance (std) of results: NeSIG obtains similar mean consistency, diversity and difficulty across 5 random seeds (STD is across seeds, not across problems) -> training is stable

Table \ref{table:experiments} compares the problems generated by NeSIG, its ablations and the domain-specific generators (\textit{ad hoc} models) using the same problem size for training and testing. 
It can be observed that NeSIG successfully learns to generate problems according to the user-defined consistency rules,
as it seldom generates inconsistent problems in \textit{blocksworld} and \textit{logistics}, and actually achieves perfect consistency (100\%) in \textit{sokoban}.
Additionally, NeSIG generates problems that are significantly more difficult than those from domain-specific generators: 3.9 times in \textit{blocksworld}, 4.75 times in \textit{logistics}, and 200 times in \textit{sokoban},
for a total (geometric) average of 15.5 times more difficulty.
Despite this,
NeSIG achieves only $8\%$ less diversity than the domain-specific generators on geometric average,
%Despite this, problems generated by NeSIG and domain-specific generators exhibit nearly equal diversity,
which is surprising considering that the random procedure followed by the latter results in highly diverse problems.
This is a remarkable result, as it means that our proposed method does not need to sacrifice diversity in order to increase problem difficulty, e.g., by learning to only generate a certain type of problem, thus effectively learning to balance difficulty and diversity.
Moreover, by leveraging parallel GPU computation, we can generate 100 problems with NeSIG in only half a minute for \textit{blocksworld} and \textit{logistics}, and in less than four minutes for \textit{sokoban}.
Finally, we observe that NeSIG consistently achieves similar results across runs, as indicated by the low standard deviations.

% NeSIG vs ablations 
% - ablations -> analizar qué es más importante, si entrenar la init o goal policy
% random vs ad hoc -> is random-both a side contribution or not?

We now turn our attention to the ablation models. 
It can be observed that using a random policy for initial state generation (\textit{random-init} and \textit{random-both} models) severely degrades consistency in \textit{blocksworld} and \textit{logistics}.
This shows that, for domains with complex consistency rules such as \textit{blocksworld} and \textit{logistics},
a trained (i.e., non-random) generative policy is needed to reliably generate consistent initial states.
%initial states generated at random are very likely to be inconsistent.
%Therefore, it is of utter importance to train the model to generate problems that are consistent, and not only diverse and hard to solve.
Additionally, ablations also significantly impair problem difficulty, although the effect of each policy ablation depends on the particular domain considered.
In \textit{blocksworld}, it is more important to train the goal policy than the initial state policy, since \textit{random-init} achieves better difficulty than \textit{random-goal}.
In \textit{sokoban}, the opposite case happens, as \textit{random-goal} achieves several orders of magnitude better difficulty than \textit{random-init}.
Finally, in \textit{logistics} the two policies seem to be equally important, as both ablations attain similar difficulty.

The \textit{random-both} model represents the full ablation where no policy is trained, thus obtaining the worst results among all models.
However, an important advantage of this model over NeSIG and the other ablations is that it does not require any type of training so, as long as consistency rules are provided (which can be done easily using our proposed consistency language),
it can be quickly applied to generate problems for any (typed-STRIPS) planning domain.
Although problems generated with this approach are easier to solve than those from the \textit{ad hoc} models,
problem difficulty can often be easily raised by incrementing problem size, just as Autoscale and \textit{ad hoc} models do.
Therefore, for cases where increasing problem size is acceptable or problem difficulty is not a concern,
the \textit{random-both} model offers a general and low-effort alternative to domain-specific generators, serving as a side contribution of our work.

% > Generalization results
% Mencionar planner generalization (usamos un conjunto distintos de planners en train y test)

Figure \ref{fig:generalization_plots} shows the difficulty obtained by NeSIG when tested on problems of different size than those used during training. We also plot the difficulty of domain-specific generators for comparison purposes.
In \textit{logistics}, NeSIG successfully generalizes to problems up to twice the size of those seen in training,
beating the \textit{ad hoc} model in terms of difficulty.
However, for sizes 35 and 40, there is a sudden spike in the difficulty of problems from the \textit{ad hoc} model, which manages to outperform NeSIG.
Our hypothesis is that the patterns learned by NeSIG about which problem features result in high difficulty do not apply to problems with more than 30 atoms.
Therefore, in order for our method to generalize past this point, it should be trained on larger problems.
In \textit{blocksworld} and \textit{sokoban},
NeSIG displays even better generalization abilities,
obtaining several times more difficulty (note the logarithmic Y-axis in Figure \ref{fig:generalization_plots}) than the domain-specific generators for every problem size tested.

% > Conclusions
% - Hablar de que los instance generators usan extensive, domain-specific prior knowledge y poner ejemplos
% - Por este motivo, los ad hoc models necesitan mucho human effort, mucho más que NeSIG (al usar nuestro semi-declarative language for consistency).

In conclusion, 
NeSIG is able to generate consistent problems with high difficulty and diversity,
successfully generalizing to problems several times larger than those seen during training.
These are remarkable results, especially taking into consideration that our method is domain-independent, whereas \textit{ad hoc} models have been tailored to each particular domain and leverage extensive domain knowledge.
For example, the \textit{blocksworld} generator uses an ad hoc formula to make sure that every consistent state has the same probability of being generated.
In \textit{logistics}, the \textit{ad hoc} model obtains the goal by randomly shuffling the packages in the initial state, knowing in advance that such a goal will always be achievable.
In \textit{sokoban}, the original ad hoc generator employed a complex procedure that allowed for little flexibility (e.g., problems of size 5x5 could not have more than two boxes).
Our new \textit{sokoban} generator does not have this limitation but, in exchange,
it is very slow (e.g., it needs 1019 seconds to generate 100 problems of size 5x5, as shown in Table \ref{table:experiments}).
%it may generate unsolvable problems.
%This is the reason why it is so slow (e.g., it needs 1019 seconds to generate 100 problems of size 5x5, as shown in Table \ref{table:experiments}),
%as it wastes much time discarding unsolvable problems.
When compared to ad hoc generators,
NeSIG requires little prior knowledge,
as it only receives as inputs the maximum problem size, the types and predicates that can appear in goals, and the set of properties (consistency constraints) that initial states must satisfy.
Moreover, with our proposed semi-declarative language,
these consistency constraints can be easily and intuitively encoded (see Appendix for concrete examples), thus reducing human effort even further.

\section{Conclusion}

% - NeSIG  NeSIG optimizes difficulty in a clever way, as it automatically learns domain-specific information (i.e., what makes a hard problem for a particular domain) and exploits it to generate challenging problems given a limit on their size.
% - Decir que hemos obtenido un SOTA method for problem generation. We hope to have shown that problems (such as those used for the IPC) should not be generated using handcrafted, domain-specific generators but, rather, with a general, domain-independent method that can learn to generate problems according to user preferences, as the one proposed in this work.

% Future work -> NeSIG can, in principle, be used to generate problems according to any type of user preferences: soft constraints encoded in reward and hard constraints encoded in consistency rules. Great flexibility.

In this work we introduced NeSIG, to the best of our knowledge the first domain-independent method for the automatic generation of planning problems that are simultaneously valid, diverse and difficult to solve.
We formulated problem generation as an MDP, training two policies with Deep RL to generate problems with the desired properties.
Both policies were encoded by NLMs, a neuro-symbolic deep neural network architecture capable of working with FOL data.

A remarkable feature of our method is that it does not require a training dataset of example problems.
Instead, it only receives as inputs the PDDL domain description and a set of consistency constraints generated problems must satisfy, along with some extra information (maximum problem size and the types and predicates that are allowed in goals).
Therefore, NeSIG requires less prior knowledge than handcrafted, domain-specific generators such as those often used in the IPCs.
Moreover, we proposed a semi-declarative language for encoding consistency constraints in an intuitive and interpretable manner,
further reducing human effort.

We tested NeSIG on three classical domains, comparing our approach against domain-specific generators and several ablations.
Results show NeSIG successfully generates valid problems which are as diverse as those from domain-specific generators but considerably more difficult (15.5 times more on geometric average).
Additionally, it showcases impressive generalization abilities,
as it generates harder problems than the ad hoc generator in \textit{logistics} for up to twice the training size,
and surpasses the difficulty of domain-specific generators in \textit{blocksworld} and \textit{sokoban} for all test sizes considered.
%as it generally obtains harder problems than the domain-specific generators when tested on larger problems than those used during training.
%as it generates harder problems than the domain-specific generators in all the domains considered when tested on larger problems than those used during training.
In light of the results obtained, we believe our work establishes a new state of the art in planning problem generation and hope it will prove useful to the Automated Planning community.

%It is important to note that the choice of consistency constraints and quality metric to optimize are totally arbitrary,
We note the choice of consistency constraints and quality metric to optimize depends on user preferences regarding the type of problems to generate.
Therefore, in future work, we plan to harness the flexibility of NeSIG by generating problems according to different user preferences (e.g., maximizing plan length instead of planning difficulty), where hard constraints will be represented as consistency rules and soft constraints numerically as part of the reward function.
We will also extend the expressivity of our method beyond typed STRIPS, e.g., by generating PDDL2.1 \cite{fox2003pddl2} problems with numeric fluents.
Additionally, we also plan to explore several applications of our method. A few examples would be \textit{automated curriculum generation}, i.e., generating problems of just the right difficulty for efficiently training an AI agent to solve a particular set of tasks,
and \textit{adversarial problem generation}, i.e.,
exploring the weaknesses of planning algorithms by generating problems that are challenging for a particular planner.
%that are difficulty for some particular planner but relatively easy for the rest.

\begin{ack}
This work has been partially funded by the Grant PID2022-142976OB-I00, funded by MICIU/AEI/ 10.13039/501100011033 and by “ERDF/EU”, as well as the Andalusian Regional predoctoral grant no. 21-111- PREDOC-0039 and by “ESF Investing in your future”.

We want to express our deep gratitude to Masataro Asai, for his suggestion to use NLMs in our work; Simon Stahlberg, for providing the implementation of Graph Neural Networks used in a previous version of this work; Mauro Vallati and the rest of authors of \cite{fawcett2014improved}, in addition to Sergio Jimenez Celorrio, for their advice on how to measure problem difficulty; Jiayuan Mao and the rest of authors of \cite{dong2019neural}, for their helpful advice on NLMs; and, finally, Christian Muise and the FastDownward (FD) community, for their invaluable help on the use of the FD planning system.
\end{ack}

\bibliography{ecai24}

% ----------- Appendices

\clearpage

\twocolumn[
  \begin{@twocolumnfalse}
    \begin{center}
      \huge \textbf{Appendix}
      \vspace{1cm} % Adjust the vertical space as needed
    \end{center}
  \end{@twocolumnfalse}
]

\appendix
% Appendix tables indices start from 1 again
\setcounter{table}{0}

\section{Problem generation example}

%\textcolor{red}{El appendix tiene que ser un PDF aparte!}

In this appendix we provide a simple, handcrafted example of our problem generation method that illustrates how a single planning problem is created from start to finish. For this example, we will use \textit{blocksworld} as our domain and, at each step (state), we will assume a random action is chosen from the set of applicable actions $A_{app}$. In the initial state generation phase, an action $a \in A_{app}$ corresponds to adding an atom to the (current) initial state $s_{ic}$, where this atom is obtained by instantiating a domain predicate ($on$, $ontable$, $handempty$, $holding$ and $clear$ in blocksworld) on objects of the correct type ($block$ type in blocksworld, as all predicates are only instantiated on objects of this type) so that $s_{ic}$ remains continuous-consistent. These objects can be in $s_{ic}$ or not. In the latter case, we call them $virtual$ objects and they are added to $s_{ic}$ along with their corresponding atom. In the goal generation phase, an action $a \in A_{app}$ corresponds to executing a domain action ($stack$, $unstack$, $pick-up$ and $put-down$ in blocksworld) in the (current) goal state $s_{gc}$, modifying the atoms in $s_{gc}$ according to the \textit{add} and \textit{delete} effects of $a$, which are encoded in the PDDL description of the domain. The selected action $a$ must be grounded, i.e., instantiated, on objects of the correct type ($block$ type in blocksworld, as all actions are applied to objects of this type) present in $s_{gc}$ and, also, its preconditions must be true in $s_{gc}$.

Additionally, we will assume we randomly choose when to stop generating the states $s_{ic}$ and $s_{gc}$. In a real scenario, the generative policies would be in charge of both selecting the next action to execute and when to stop generating $s_{ic}$ and $s_{gc}$ (i.e., sampling the termination action $end$). We represent the states of the MDP, corresponding to (incomplete or fully-generated) planning problems, as a tuple $(s_{ic}, s_{gc})$. We represent $s_{ic}$ and $s_{gc}$ as another tuple $(O, P)$, where $O$ is a set containing the objects present in the state (with their respective types), and $P$ is a set containing the atoms of the state. We now detail the process followed to generate the example problem:

\begin{enumerate}
    \item At the start of the generation process, the goal state $s_{gc}$ is empty and the initial state $s_{ic}$ can be either empty or equal to another state predefined by the user. In blocksworld, $s_{ic}$ also begins empty. Therefore, the generation process starts from an empty MDP state $(\_, \_)$, where the first element of the tuple represents $s_{ic}$ and the second one represents $s_{gc}$.
    We assume the action $add \: ontable(o1)$ is sampled, where $o1$ is an object of type $block$.
    Since $o1$ corresponds to a virtual object, we also need to add it to $s_{ic}$. Thus, the resulting state is $( \: (\{block \; o1\}, \, \{ontable(o1)\}), \: \_)$, which corresponds to a continuous-consistent state.
    
    \item The action $add \: on(o2,o1)$ is selected, where $o2$ is a virtual object of type $block$. The resulting state is $( \: (\{block \; o1, block \; o2\}, \, \{ontable(o1), on(o2,o1)\}), \: \_)$, which corresponds to a continuous-consistent state.
    
    \item The action $add \: clear(o2)$ is selected and the resulting state is $( \: (\{block \; o1, block \; o2\}, \, \{ontable(o1), on(o2,o1), \\ clear(o2)\}), \: \_)$, which corresponds to a continuous-consistent state.

    \item The action $add \: handempty()$ is selected and the resulting state is $( \: (\{block \; o1, block \; o2\}, \, \{ontable(o1), on(o2,o1), \\ clear(o2), handempty()\}), \: \_)$, which corresponds to a continuous-consistent state.

    \item The termination action $end$ is sampled, so the initial state generation phase concludes, i.e., $s_i=s_{ic}$. Then, the consistency evaluator checks if $s_i$ meets the eventual consistency rules, which it does (otherwise, $s_i$ would be discarded and no goal would be generated). Therefore, the goal generation phase can start. The goal state is initialized to the initial state, i.e., $s_{gc}=s_i$, so the resulting state is $( \: (\{block \; o1, block \; o2\}, \, \{ontable(o1), on(o2,o1), \\ clear(o2), handempty()\}), (\{block \; o1, block \; o2\}, \\ \{ontable(o1),  on(o2,o1), clear(o2), handempty()\}))$.
   
    % \item The action $add \: handempty() $ is selected and the resulting state is continuous-consistent. Assume the initial state generation phase concludes at this step. If that is the case, the initial state has been completely generated, i.e., $s_i=s_{ic}$, and the goal generation phase starts from $s_i$, i.e., $s_{gc} = s_i$. Thus, the next state is actually $s_5 = ( \: (\{block \; o1, block \; o2\}, \, \{ontable(o1), on(o2,o1), \\ clear(o2), handempty()\}), (\{block \; o1, block \; o2\}, \\ \{ontable(o1),  on(o2,o1), clear(o2), handempty()\}))$.
    
    \item As we are now in the goal generation phase, the set of applicable actions $A_{app}$ corresponds to the domain actions whose preconditions are met in $s_{gc}$. Assume the action $unstack(o2, o1)$ is selected. Then, the current goal state $s_{gc}$ is modified with the effects of the selected action. Thus, the next state is $( \: (\{block \; o1, block \; o2\}, \, \{ontable(o1), on(o2,o1), \\ clear(o2), handempty()\}),
    (\{block \; o1, block \; o2\}, \\ \{ontable(o1), holding(o2), clear(o1)\}))$.
    
    \item The action $put-down(o2)$ is selected and the resulting state is $( \: (\{block \; o1, block \; o2\}, \, \{ontable(o1), on(o2,o1), \\ clear(o2), handempty()\}),
    (\{block \; o1, block \; o2\}, \\ \{ontable(o1), clear(o1), clear(o2), handempty(), \\ ontable(o2)\}))$.
    
    \item The termination action $end$ is sampled, so the goal generation phase concludes, i.e., $s_g = s_{gc}$. Then, the problem goal $g$ is obtained by selecting a subset of the atoms in $s_g$ according to the goal types and predicates given by the user. Let us assume the goal types and predicates list is $\{ontable(block), on(block,block)\}$, i.e., $g$ must only contain atoms of predicate type $on$ or $ontable$ instantiated on objects of type $block$. If that is the case, the goal is $g=\{ontable(o1), ontable(o2)\}$ and NeSIG outputs the problem $(s_i, g)$, which is shown in Listing \ref{lst:listing_1}.
\end{enumerate}

\begin{lstlisting}[caption=Example problem generated with NeSIG., label={lst:listing_1}]
(define (problem example_blocksworld_problem)

(:domain blocksworld)

(:objects o1 o2 - block)

(:init (ontable o1) (on o2 o1) 
       (clear o2) (handempty))

(:goal (ontable o1) (ontable o2))
)
\end{lstlisting}

\section{PDDL domains}

In this appendix we provide the PDDL descriptions for the three planning domains employed in our experiments: \textit{blocksworld}, \textit{logistics} and \textit{sokoban}.

\begin{lstlisting}[caption=PDDL description for \textit{blocksworld}., label={lst:pddl_blocksworld}]
(define (domain BLOCKS)
  (:requirements :strips :typing)
  (:types block)
  (:predicates (on ?x - block ?y - block)
               (ontable ?x - block)
               (clear ?x - block)
               (handempty)
               (holding ?x - block))

  (:action pick-up
          :parameters (?x - block)
          :precondition (and (clear ?x) 
                             (ontable ?x) 
                             (handempty))
	     :effect (and (not (ontable ?x))
                       (not (clear ?x))
                       (not (handempty))
                       (holding ?x)))

  (:action put-down
          :parameters (?x - block)
          :precondition (holding ?x)
          :effect (and (not (holding ?x))
                       (clear ?x)
                       (handempty)
                       (ontable ?x)))
 
  (:action stack
          :parameters (?x - block ?y - block)
          :precondition (and (holding ?x) 
                             (clear ?y))
          :effect (and (not (holding ?x))
                       (not (clear ?y))
                       (clear ?x)
                       (handempty)
                       (on ?x ?y)))

  (:action unstack
          :parameters (?x - block ?y - block)
          :precondition (and (on ?x ?y) 
                             (clear ?x) 
                             (handempty))
          :effect (and (holding ?x)
                       (clear ?y)
                       (not (clear ?x))
                       (not (handempty))
                       (not (on ?x ?y)))))
\end{lstlisting}

\begin{lstlisting}[caption=PDDL description for \textit{logistics}., label={lst:pddl_logistics}]
(define (domain logistics)
  (:requirements :strips :typing :existential-preconditions) 
  (:types city location thing - object
          package vehicle - thing
          truck airplane - vehicle  
          airport - location)
  (:predicates (in-city ?l - location ?c - city)
               (at ?obj - thing ?l - location)
               (in ?p - package ?veh - vehicle))
  
  (:action drive
          :parameters (?t - truck ?from ?to - location)
          :precondition (and (at ?t ?from)
                             (exists (?c - city)
                             (and (in-city ?from ?c) 
                                  (in-city ?to ?c))))
          :effect (and (not (at ?t ?from))
                       (at ?t ?to)))

  (:action fly
          :parameters (?a - airplane ?from ?to - airport)
          :precondition (at ?a ?from)
          :effect (and (not (at ?a ?from))
                       (at ?a ?to)))

  (:action load
          :parameters (?v - vehicle ?p - package ?l - location)
          :precondition (and (at ?v ?l)
                        (at ?p ?l))
          :effect (and (not (at ?p ?l))
                       (in ?p ?v)))

  (:action unload
          :parameters (?v - vehicle ?p - package ?l - location)
          :precondition (and (at ?v ?l)
                             (in ?p ?v))
          :effect (and (not (in ?p ?v))
                       (at ?p ?l))))
\end{lstlisting}

\begin{lstlisting}[caption=PDDL description for \textit{sokoban}., label={lst:pddl_sokoban}]
(define (domain sokoban)
  (:requirements :typing :negative-preconditions)
  (:types loc)
  (:predicates (at-robot ?l - loc)
               (at-box ?l - loc)
               (at-wall ?l - loc)
               (connected-up ?l1 - loc ?l2 - loc)
               (connected-right ?l1 - loc ?l2 - loc))
               
  (:action move-up
          :parameters (?from - loc ?to - loc)
          :precondition (and (at-robot ?from)
                             (not (at-box ?to))
                             (not (at-wall ?to))
                             (connected-up ?from ?to))
          :effect (and (not (at-robot ?from))
                       (at-robot ?to)))

  (:action move-down
          :parameters (?from - loc ?to - loc)
          :precondition (and (at-robot ?from)
                             (not (at-box ?to))
                             (not (at-wall ?to))
                             (connected-up ?to ?from))
          :effect (and (not (at-robot ?from))
                       (at-robot ?to)))

  (:action move-right
          :parameters (?from - loc ?to - loc)
          :precondition (and (at-robot ?from)
                             (not (at-box ?to))
                             (not (at-wall ?to))
                             (connected-right ?from ?to))
          :effect (and (not (at-robot ?from))
                       (at-robot ?to)))

  (:action move-left
          :parameters (?from - loc ?to - loc)
          :precondition (and (at-robot ?from)
                             (not (at-box ?to))
                             (not (at-wall ?to))
                             (connected-right ?to ?from))
          :effect (and (not (at-robot ?from))
                       (at-robot ?to)))

  (:action push-up
          :parameters (?l1 - loc ?l2 - loc ?l3 - loc)
          :precondition (and (at-robot ?l1)
                             (at-box ?l2)
                             (not (at-box ?l3))
                             (not (at-wall ?l3))
                             (connected-up ?l1 ?l2)
                             (connected-up ?l2 ?l3))
          :effect (and (not (at-robot ?l1))
                       (at-robot ?l2)
                       (not (at-box ?l2))
                       (at-box ?l3)))

  (:action push-down
          :parameters (?l1 - loc ?l2 - loc ?l3 - loc)
          :precondition (and (at-robot ?l1)
                             (at-box ?l2)
                             (not (at-box ?l3))
                             (not (at-wall ?l3))
                             (connected-up ?l2 ?l1)
                             (connected-up ?l3 ?l2))
          :effect (and (not (at-robot ?l1))
                       (at-robot ?l2)
                       (not (at-box ?l2))
                       (at-box ?l3)))

  (:action push-right
          :parameters (?l1 - loc ?l2 - loc ?l3 - loc)
          :precondition (and (at-robot ?l1)
                             (at-box ?l2)
                             (not (at-box ?l3))
                             (not (at-wall ?l3))
                             (connected-right ?l1 ?l2)
                             (connected-right ?l2 ?l3))
          :effect (and (not (at-robot ?l1))
                       (at-robot ?l2)
                       (not (at-box ?l2))
                       (at-box ?l3)))

  (:action push-left
          :parameters (?l1 - loc ?l2 - loc ?l3 - loc)
          :precondition (and (at-robot ?l1)
                             (at-box ?l2)
                             (not (at-box ?l3))
                             (not (at-wall ?l3))
                             (connected-right ?l2 ?l1)
                             (connected-right ?l3 ?l2))
          :effect (and (not (at-robot ?l1))
                       (at-robot ?l2)
                       (not (at-box ?l2))
                       (at-box ?l3))))
\end{lstlisting}

\section{Consistency rules}

In this appendix we detail the consistency rules (i.e., constraints) used for the \textit{blocksworld}, \textit{logistics} and \textit{sokoban} domains. As explained in Section \ref{section:validity} of the main paper, consistency rules are encoded using a novel, semi-declarative language that makes possible to combine standard Python code with a FOL-like syntax. For each domain, we need to provide a consistency evaluator containing the consistency rules, which simply corresponds to a Python class with two methods: \textit{check\_continuous\_consistency(self, curr\_state, atom\_pred, atom\_obj\_consts, atom\_obj\_inds, atom\_obj\_types)} and \textit{check\_eventual\_consistency(self, curr\_state)}.
The first method receives as inputs the current initial state $s_{ic}$ (\textit{curr\_state}) along with some atom $a$ to add to $s_{ic}$, whose information is encoded in the \textit{atom\_pred}, \textit{atom\_obj\_consts}, \textit{atom\_obj\_inds} and \textit{atom\_obj\_types} parameters, and returns whether the state resulting from adding $a$ to $s_{ic}$ is continuous-consistent or not. 
Both \textit{atom\_obj\_consts} and \textit{atom\_obj\_inds} contain the objects instantiated on the atom $a$ but represent this information in a different manner: the former is used by rules encoded as FOL formulas, whereas the latter is employed by rules encoded using standard Python code. Therefore, the user can encode consistency rules in their preferred manner: declarative (i.e., FOL) or imperative (i.e., Python).
This choice is completely transparent to NeSIG and does not impact training.
The second method receives as its sole input the completely-generated initial state $s_i$ and returns whether it is eventual-consistent or not.
%We now provide the complete and precise code, including comments, for the consistency evaluators of the three domains employed in our work.
We now provide the code for the consistency evaluators of the three domains employed in our work.
For brevity reasons, we have omitted lines corresponding to aliases (e.g., \textit{ontable = self.ontable}) and declaration of FOL variables (e.g., \textit{x = Variable('x')}).
As previously mentioned, consistency rules are chosen so that NeSIG can generate the same set of problems as the domain-specific generators.
For instance, problems obtained by the blocksworld generator never contain atoms of type $holding$ in their initial state, 
so this constraint is also encoded in the consistency rules for blocksworld employed by NeSIG.
Finally, it can be observed that, thanks to our proposed language,
consistency rules can be encoded in an easy and interpretable manner with just a few lines of code.

\begin{lstlisting}[style=pythoncode, caption=Consistency evaluator for \textit{blocksworld}., label={lst:evaluator_blocksworld}]
class ConsistencyEvaluatorBlocksworld(ConsistencyEvaluator):

    def check_continuous_consistency(self, 
    curr_state, atom_pred, atom_obj_consts, 
    atom_obj_inds, atom_obj_types):

        """
        (ontable x)
            - x is new
        """
        if atom_pred == 'ontable':
            a = atom_obj_consts[0]
            formula = virtual(a)
            return self._evaluate(formula)

        """
        (on a b)
            - a is new
            - b is NOT new
            - b does not appear in an atom of type 
              clear
            - b does not have a block on top of 
              it (on(*,b) does not exist)
        """
        if atom_pred == 'on':
            a, b = atom_obj_consts
            formula = virtual(a) & ~virtual(b) & ~clear(b) & ~TE(x, on(x,b))
            return self._evaluate(formula)

        """
        (clear x)
            - x is NOT new
            - x does not appear in holding(x)
            - x does not have a block on top of
              it (on(*,x) does not exist)
        """
        if atom_pred == 'clear':
            a = atom_obj_consts[0]
            formula = ~virtual(a) & ~TE(x, on(x,a))
            return self._evaluate(formula)

        """
        (holding x)
            - The initial state cannot contain
              atoms of type "holding"
        """
        if atom_pred == 'holding':
            return False

        """
        (handempty)
            No consistency rules to check
            Note that we don't need to check
            whether (handempty) already exists,
            since repeated atoms are implicitly
            discarded
        """
        if atom_pred == 'handempty':
            return True

    def check_eventual_consistency(self, curr_state):
        # The problem must contain at least 
        # two blocks (otherwise, they can't be 
        # stacked)
        formula_1 = (TE(x, _type(x, block)) >= 2)

        # The initial state must contain 
        # (handempty)
        formula_2 = handempty()

        # For all objects x there must exist 
        # clear(x), unless they have another 
        # block y on top
        formula_3 = FA(x,clear(x) | TE(y, on(y,x)))
        
        return self._evaluate(formula_1 & formula_2 
        & formula_3)
\end{lstlisting}

\begin{lstlisting}[style=pythoncode, caption=Consistency evaluator for \textit{logistics}., label={lst:evaluator_logistics}]
class ConsistencyEvaluatorLogistics(ConsistencyEvaluator):

    def check_continuous_consistency(self, 
    curr_state, atom_pred, atom_obj_consts,
    atom_obj_inds, atom_obj_types):
        """
        (in-city ?loc - location ?city - city)
            - loc is new
            - if city is new, then loc must be of 
              type airport (the first location of 
              every city is always an airport)
            - if city is NOT new, then loc must 
              be of type location (since each 
              city contains one and only one 
              airport)
        """
        if atom_pred == 'in-city':
            loc, city = atom_obj_consts
            formula = virtual(loc) & 
            (_type(loc,airport) ** virtual(city))
            return self._evaluate(formula)

        """
        (at ?obj - thing ?loc - location)
            - obj is new
            - loc must not be new
            - obj must be of type package, truck or 
              airplane
            - if obj is of type airplane, then loc 
              must be of type airport
        """
        if atom_pred == 'at':
            obj, loc = atom_obj_consts
            formula = virtual(obj) & ~virtual(loc)
            & ( _type(obj, package) | 
            _type(obj, truck) | 
            _type(obj, airplane) ) & 
            ( _type(obj, airplane) >> 
            _type(loc, airport) )
            return self._evaluate(formula)

        """
        (in ?p - package ?veh - vehicle)
            The initial state can have no atoms of 
            type "in"
        """
        if atom_pred == 'in':
            return False

    def check_eventual_consistency(self, curr_state):
        # The problem must contain at least one 
        # airplane
        formula_1 = TE(x, _type(x, airplane))
 
        # The problem must contain at least two
        # cities
        formula_2 = TE(x, _type(x, city)) >= 2

        # Every city must contain at least one 
        # truck
        # x -> city, y -> location/airport in 
        # the city, z -> truck at the 
        # location/airport
        # Meaning of the formula: "For every 
        # city x, there must exist a 
        # location/airport y in the city x, 
        # so that there exists a truck z at 
        # the location/airport y"
        formula_3 = FA(x, _type(x, city) >> 
        TE(y, in_city(y, x) & 
        TE(z, _type(z, truck) & at(z, y) ) ) )
        
        return self._evaluate(formula_1 & formula_2
        & formula_3)
\end{lstlisting}

\begin{lstlisting}[style=pythoncode, caption=Consistency evaluator for \textit{sokoban}., label={lst:evaluator_sokoban}]
class ConsistencyEvaluatorSokoban(ConsistencyEvaluator):

    def check_continuous_consistency(self, curr_state, atom_pred, atom_obj_consts, atom_obj_inds, atom_obj_types):
        """
        (connected-up l1 l2)
            - Cannot be added, as the we start 
              generation from an empty NxM map
        """
        if atom_pred == 'connected-up':
            return False

        """
        (connected-right l1 l2)
            - Cannot be added, as the we start 
              generation from an empty NxM map
        """
        if atom_pred == 'connected-right':
            return False 

        """
        (at-robot loc)
            - loc must already exist in the state
            - Only one robot can exist at the same 
              time
            - (at-box loc) does not exist
            - (at-wall loc) does not exist
        """
        if atom_pred == 'at-robot':
            loc = atom_obj_consts[0]
            formula = ~virtual(loc) & 
            ~TE(x, at_robot(x)) & ~at_box(loc) 
            & ~at_wall(loc)
            return self._evaluate(formula)

        """
        (at-box loc)
            - loc must already exist in the state
            - (at-robot loc) does not exist
            - (at-wall loc) does not exist
        """
        if atom_pred == 'at-box':
            loc = atom_obj_consts[0]
            formula = ~virtual(loc) & 
            ~at_robot(loc) & ~at_wall(loc)
            return self._evaluate(formula)

        """
        (at-wall loc)
            - loc must already exist in the state
            - (at-robot loc) does not exist
            - (at-box loc) does not exist
        """
        if atom_pred == 'at-wall':
            loc = atom_obj_consts[0]
            formula = ~virtual(loc) & 
            ~at_robot(loc) & ~at_box(loc)
            return self._evaluate(formula)

    def check_eventual_consistency(self, curr_state):
        # The initial state must contain one robot
        formula = (TE(x, at_robot(x)) == 1)
        
        return self._evaluate(formula)
\end{lstlisting}

\section{Parameters for domain-specific generators}
In this appendix we detail the range of parameters employed for the domain-specific generator of each domain: \textit{blocksworld}, \textit{logistics} and \textit{sokoban}.
For each generator parameter, we set a possible range of values and uniformly sample from it to generate each problem.
In blocksworld and logistics, if the size (i.e., number of atoms in the initial state) of the generated problem is smaller than $D-2$, where $D$ is the maximum problem size for the particular experiment, we discard the problem and retry generation.
This is done in order to provide a fair comparison with NeSIG, since smaller problems tend to be easier to solve,
and the time spent discarding small problems is not considered when measuring generation time.
In sokoban, instead, we make sure that the problem obtained contains at least a 25\% of empty cells (with no walls or boxes) since, otherwise, the problem is very likely to be unsolvable.
Finally, the number of problems generated per experiment, time and memory limits for the planner, and difficulty of terminated problems is equal to that used for NeSIG (see Table \ref{table:hyperparameters}). The remaining configuration parameters are detailed in Tables \ref{table:bw_generator_params}, \ref{table:lg_generator_params} and \ref{table:sk_generator_params}.
We have selected these values in order to maximize problem diversity and avoid generation bias towards any particular type of problem.

\begin{table}
\caption{\textbf{Generator parameters for \textit{blocksworld}.} $D$ denotes maximum problem size and $ceil$ is the ceiling function that approximates a real number to the next integer.}
\label{table:bw_generator_params}
\begin{center}
\begin{tabular}{lr}
\toprule
\multicolumn{1}{c}{Name} & \multicolumn{1}{c}{Value} \\
\cmidrule{1-2}
seed for each experiment & 1 \\
blocks range & $[ceil(D/3), D]$ \\
\bottomrule
\end{tabular}
\end{center}
\end{table}

\begin{table}
\caption{\textbf{Generator parameters for \textit{logistics}.} $D$ denotes maximum problem size.}
\label{table:lg_generator_params}
\begin{center}
\begin{tabular}{lr}
\toprule
\multicolumn{1}{c}{Name} & \multicolumn{1}{c}{Value} \\
\cmidrule{1-2}
seed for each experiment & 1 \\
airplanes range & $[1,D]$ \\
cities range & $[2,D]$ \\
city\_size range & $[1,D]$ \\
packages range & $[1,D]$ \\
extra\_trucks range & $[0,D]$ \\
\bottomrule
\end{tabular}
\end{center}
\end{table}

\begin{table}
\caption{\textbf{Generator parameters for \textit{sokoban}.} $C$ denotes the number of cells (e.g., 25 for a map size of 5x5) and $floor$ is the floor function that approximates a real number to the preceding integer.}
\label{table:sk_generator_params}
\begin{center}
\begin{tabular}{lr}
\toprule
\multicolumn{1}{c}{Name} & \multicolumn{1}{c}{Value} \\
\cmidrule{1-2}
seed for each experiment & 1 \\
boxes range & $[1,floor(C*0.6)]$ \\
walls range & $[0,floor(C*0.6)]$ \\
\bottomrule
\end{tabular}
\end{center}
\end{table}

\section{NeSIG hyperparameters}
\label{appendix:hyperparameters}

In this appendix we provide a comprehensive list with all the NeSIG hyperparameters used in this work. We have utilized the same hyperparameter values for the three domains: \textit{blocksworld}, \textit{logistics} and \textit{sokoban}. The exceptions to this rule are the following: goal types and predicates\footnote{We take into consideration type inheritance. Therefore, the set $\{at(package,location)\}$ is equivalent to $\{at(package,location), at(package,airport)\}$, since type $airport$ inherits from $location$.}, since this information is domain-dependent; the maximum number of actions in the goal generation phase, as we use more for sokoban since this domain is more challenging than the others; diversity threshold $\theta$ (see Equation \ref{eq:total_reward}), chosen by testing a few values and selecting the one resulting in the best balance between problem difficulty and diversity.
The complete list of hyperparameters is detailed in Table \ref{table:hyperparameters}.

\begin{table*}[b]
\caption{\textbf{NeSIG hyperparameters.}}
\label{table:hyperparameters}
\begin{center}
\begin{tabular}{lr}
\toprule
\multicolumn{1}{c}{Name} & \multicolumn{1}{c}{Value} \\
  \cmidrule{1-2}
  seed for each experiment & 1-5 \\
  goal types and predicates &  \\
  --- blocksworld & $\{on(block,block)\}$ \\
  --- logistics & $\{at(package,location)\}$ \\
  --- sokoban & $\{at\text{-}box(loc)\}$ \\
  max actions init state generation phase (training/validation) & 15 \\
  max actions goal generation phase (training/validation) & \\
  --- blocksworld & 60 \\
  --- logistics & 60 \\
  --- sokoban & 75 \\
  max actions init state generation phase (size generalization experiments) & \\
  --- blocksworld & [10,15,20,25,30,35,40] \\
  --- logistics & [10,15,20,25,30,35,40] \\
  --- sokoban & [10,15,21,30] \\
  max actions goal generation phase (size generalization experiments) & \\
  --- blocksworld & [40,60,80,100,120,140,160] \\
  --- logistics & [40,60,80,100,120,140,160] \\
  --- sokoban & [50,75,105,150] \\
  diversity threshold $\theta$ in Equation \ref{eq:total_reward} & \\
  --- blocksworld & 0.02 \\
  --- logistics & 0.2 \\
  --- sokoban & 0.02 \\
  difficulty reward $r_f$ for problems terminated (training/validation) & $10^6$ nodes \\
  difficulty for problems terminated (test) & $10^8$ nodes \\
  planner time limit (training/validation) & 300 s \\
  planner time limit (test) & 1800 s \\
  planner memory limit (training/validation) & 512000 KB \\
  planner memory limit (test) & 8500000 KB \\ 
  training steps & 5000 \\
  learning rate & $10^{-3}$ \\
  validation period & 250 \\
  minibatch size & 64 \\
  L2 weight decay & 0.0 \\  
  reward discount factor $\gamma$ & 1.0 \\
  GAE factor $\lambda$ & 1.0 \\
  generated problems per training step & 25 \\
  generated problems per validation step & 100 \\
  generated problems per test experiment & 100 \\
  min num samples for performing a training step & 32 \\
  critic loss weight & 0.1 \\
  gradient clipping value & 5.0 \\  
  PPO epochs per train step & 3 \\
  PPO $\epsilon$ value & 0.2 \\
  Policy entropy bonus & Linearly annealed from 0.2 to 0 over 2500 training steps \\ 
  NLM & \\
  --- breadth & 3 \\
  --- depth & 5 \\
  --- output predicates for inner layers & 8 (for each arity from 0 to 3) \\
  --- MLP hidden features & 0 \\
  --- residual type & \textit{all} (concatenate to the input of each layer the outputs of all previous layers) \\
  --- exclude-self & True (\textit{reduce} operations ignore repeated tensor positions) \\
  --- batch normalization & False \\
  --- activation function & sigmoid \\
\bottomrule
\end{tabular}
\end{center}
\end{table*}

\section{Experiments with optimal planners}

\begin{figure*}[ht]
  \centering
  \includegraphics[width=0.33\textwidth]{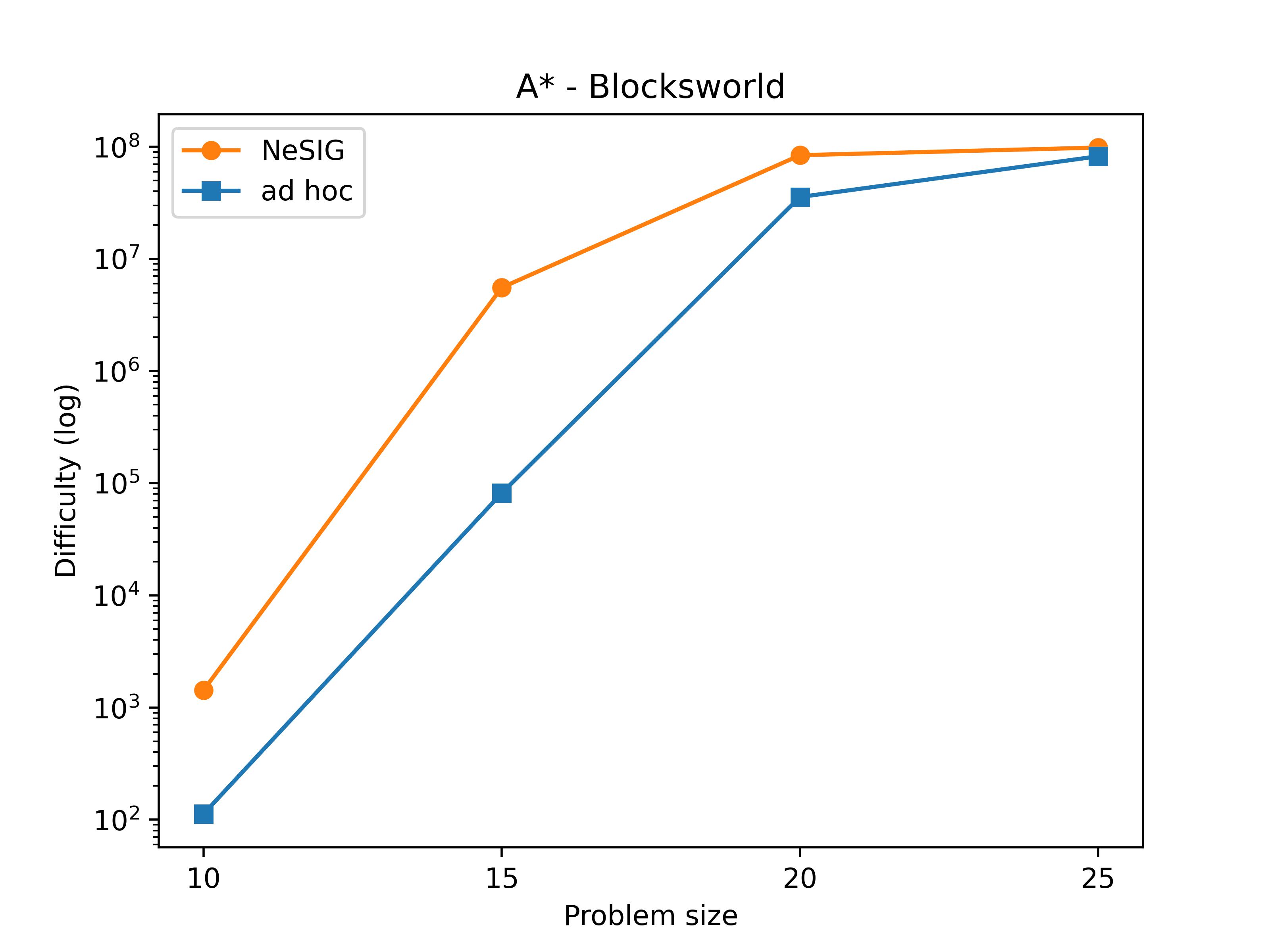}
  \includegraphics[width=0.33\textwidth]{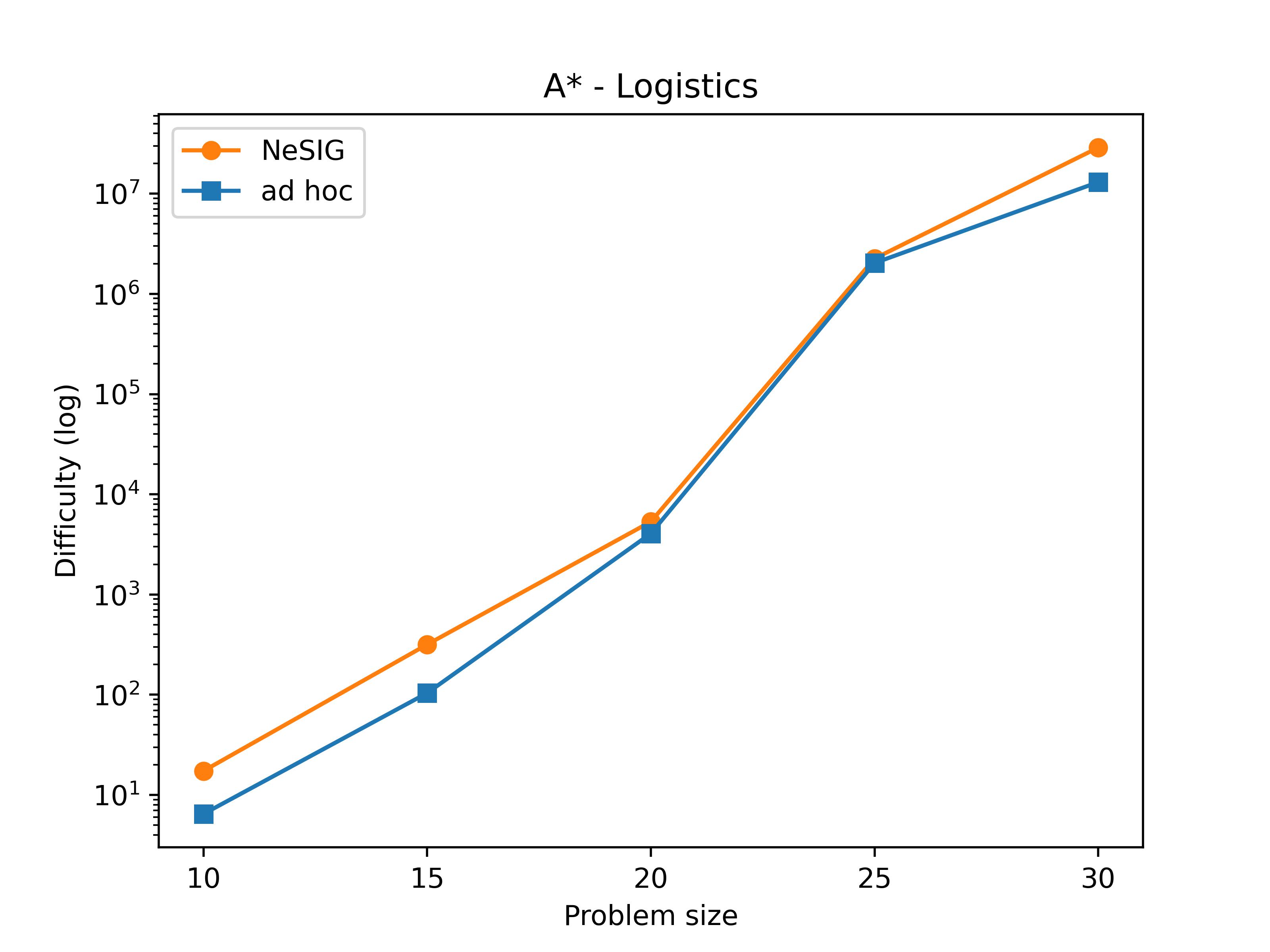}
  \includegraphics[width=0.33\textwidth]{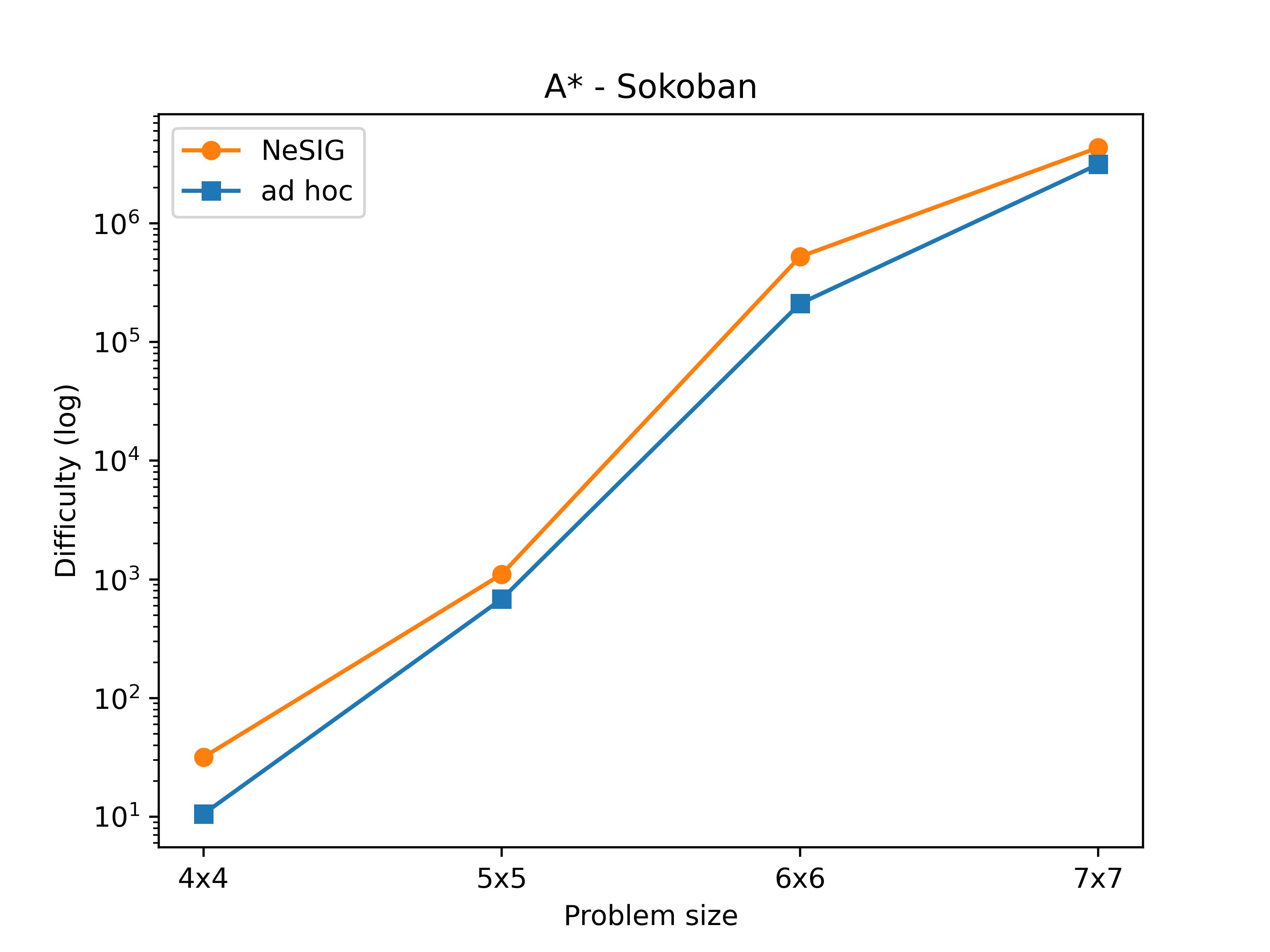}
  \caption{\textbf{Problem difficulty for A*+LM-cut.}
  The plots show the mean difficulty (in log scale), measured as the number of expanded nodes, obtained by NeSIG and \textit{ad hoc} models with the A*+LM-cut optimal planning algorithm.
  \vspace{0.2cm}
  }
  \label{fig:plots_astar}
\end{figure*}

\begin{figure*}[ht]
  \centering
  \includegraphics[width=0.33\textwidth]{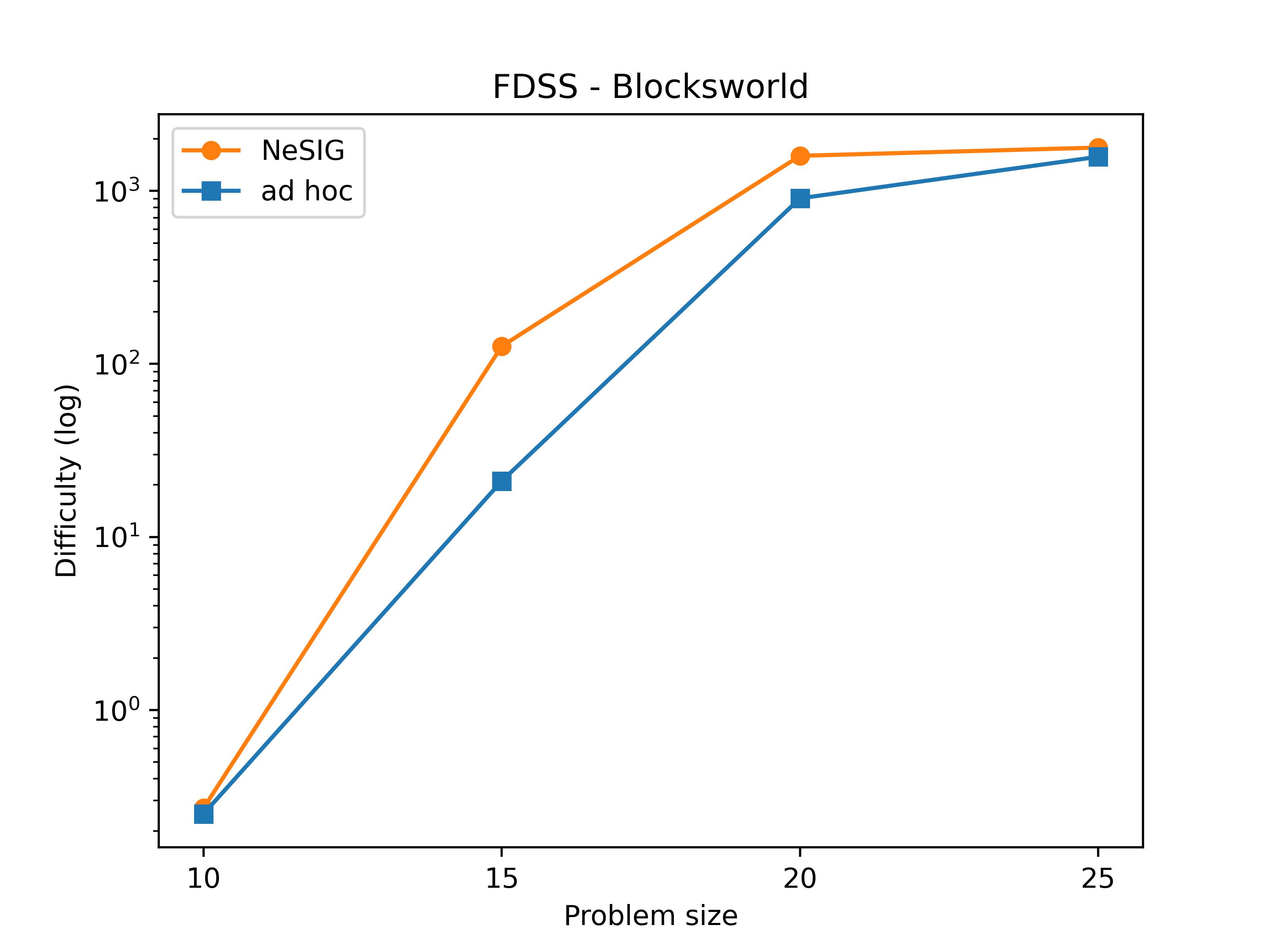}
  \includegraphics[width=0.33\textwidth]{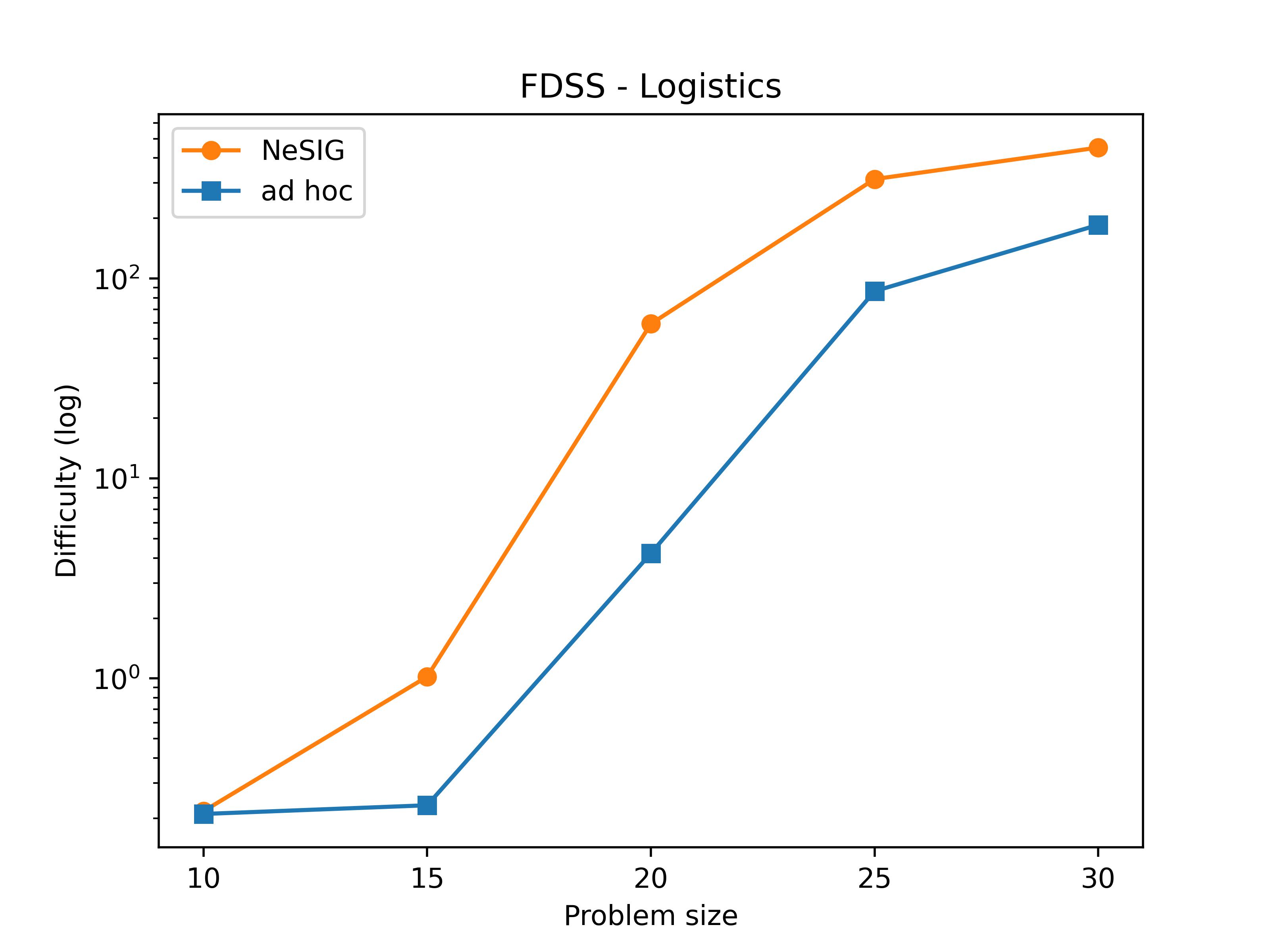}
  \includegraphics[width=0.33\textwidth]{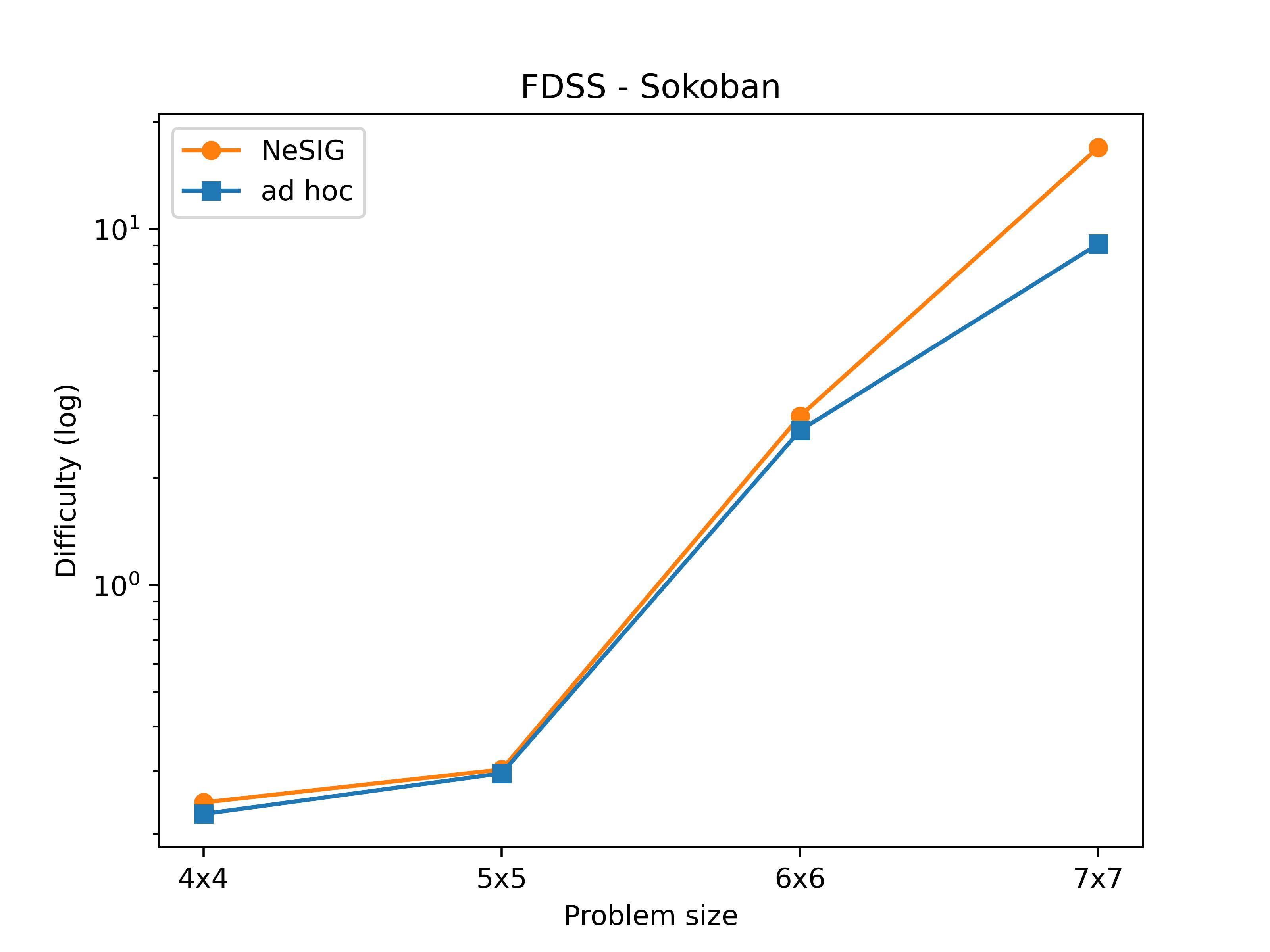}
  \caption{\textbf{Problem difficulty for FDSS.}
  The plots show the mean difficulty (in log scale), measured as the planning time in seconds, obtained by NeSIG and \textit{ad hoc} models with the FDSS optimal planning portfolio.
  \vspace{0.2cm}
  }
  \label{fig:plots_fdss}
\end{figure*}

In this appendix we extend the plots of Figure \ref{fig:generalization_plots} with the difficulties of two optimal planners:
A* \cite{hart1968formal} with the LM-cut heuristic \cite{helmert2009landmarks} (see Figure \ref{fig:plots_astar}), and the FastDownward Stone Soup (FDSS) planning portfolio \cite{helmert2011fast}, optimizing version 1 (see Figure \ref{fig:plots_fdss}).
For A*, we measured difficulty as the number of expanded nodes. We utilized the same time limit of 30 minutes and difficulty of $10^8$ for terminated problems in logistics and sokoban whereas, in blocksworld, we had to increase the time limit to one hour.
Since FDSS employs several planning algorithms, we opted for measuring difficulty as the total planning time (in seconds) needed to find an optimal plan.
We used a time limit of 10 minutes and difficulty of $601$ for terminated problems in logistics and sokoban whereas, in blocksworld, we had to increase the time limit to 30 minutes and terminated problem difficulty to $1801$.
For both planners we set a maximum memory of 8GB.
Due to their optimizing nature, these planners (especially A*) often require more planning time than the satisficing planners used in Figure \ref{fig:generalization_plots}.
Therefore, we could only test problems up to size 25 in blocksworld and 30 in logistics, whereas in sokoban we were able to test all sizes.
Despite this reduction in size, Figure \ref{fig:plots_astar} shows the A* difficulty approximates the maximum value of $10^8$ (corresponding to all-timeouts) for the largest blocksworld and logistics problems.

We observe how, despite training with a satisficing planner (LAMA-first), NeSIG successfully generalizes to optimal planners, generating more difficult problems than the \textit{ad hoc} models for every domain and problem size.
It is important to note that the Y-axis is logarithmic, so a small difference in Y-value actually translates to a much larger (exponentially larger) difference in difficulty.
Take as an example the A* difficulty for logistics problems of size 15.
Although it could seem NeSIG and the \textit{ad hoc} generator achieve similar difficulty, NeSIG actually obtains problems three times as difficult as the former.

% CAMBIAR HIPERPARÁMETROS TABLA APPENDIX

\end{document}